%% file: main.tex
\pdfoutput=1
\documentclass[11pt]{article}

\usepackage[]{acl2024}

\usepackage{times}
\usepackage{latexsym}

\usepackage[T1]{fontenc}
\usepackage{microtype}

\usepackage{inconsolata}

\usepackage[utf8]{inputenc}
\usepackage{booktabs,multirow}
\usepackage{enumitem}
\usepackage{kotex} % Menu - Tex Live version - 2021 로 설정필요
\usepackage{color}
\usepackage{rotating}
\usepackage{amsmath}
\usepackage{amsfonts}
\usepackage{pifont}
\usepackage{array}
\usepackage{amssymb}
\usepackage{algpseudocode}
\usepackage{algorithm}
\usepackage{float}
\usepackage{tcolorbox}
\usepackage{multicol}
\usepackage{lipsum}
\usepackage[normalem]{ulem}
\usepackage{xcolor}
\usepackage{soul}
\usepackage{listings}
\usepackage{xcolor}
\usepackage{amsmath} % For LaTeX math symbols
\usepackage{adjustbox}

\definecolor{customhighlight}{HTML}{DDB6E4}
\sethlcolor{customhighlight}

\colorlet{instructionbg}{gray!15}
\colorlet{questionbg}{gray!25}

\newcommand{\CommentAligned}[1]{%
  \Statex \hspace{-\algorithmicindent} \textbf{#1}%
}

\algrenewcommand\algorithmicrequire{\textbf{Input:}}
\algrenewcommand\algorithmicensure{\textbf{Output:}}

\newcommand{\ie}{{\it i.e.}}
\newcommand{\eg}{{\it e.g.}}

\title{From Reading to Compressing: Exploring the Multi-document Reader for Prompt Compression}

\author{Eunseong Choi, Sunkyung Lee, Minjin Choi, June Park, Jongwuk Lee\thanks{\ \ Corresponding author} \\
        Sungkyunkwan University, Republic of Korea\\  
        \texttt{\{eunseong, sk1027, zxcvxd, pj00515, jongwuklee\}@skku.edu}}

\begin{document}
\maketitle
\input{sec-abstract}
\input{sec-introduction}
\input{sec-related_work}

\input{sec-method}

\input{sec-exp_setup}
\input{sec-results}
\input{sec-conclusion}
\input{sec-limitation}
\input{sec-ethics}

% \bibliographystyle{acl_natbib}
\bibliography{references}

\input{sec-appendix}

\end{document}

%% file: sec-abstract.tex
\begin{abstract}

Large language models (LLMs) have achieved significant performance gains using advanced prompting techniques over various tasks. However, the increasing length of prompts leads to high computational costs and often obscures crucial information. Prompt compression has been proposed to alleviate these issues, but it faces challenges in (i) capturing the global context and (ii) training the compressor effectively. To tackle these challenges, we introduce a novel prompt compression method, namely \emph{Reading To Compressing (R2C)}, utilizing the Fusion-in-Decoder (FiD) architecture to identify the important information in the prompt. Specifically, the cross-attention scores of the FiD are used to discern essential chunks and sentences from the prompt. R2C effectively captures the global context without compromising semantic consistency while detouring the necessity of pseudo-labels for training the compressor. Empirical results show that R2C retains key contexts, enhancing the LLM performance by 6\% in out-of-domain evaluations while reducing the prompt length by 80\%. The code is available in our  \href{https://github.com/eunseongc/R2C}{repository}.

% R2C retains critical information without sacrificing performance
% and improving efficiency by up to 14.5 times compared to baseline methods.

%retaining critical information without sacrificing performance.

%Large language models have yielded significant performance gains across various tasks owing to advanced prompting techniques. However, the increasing length of prompts requires high computational costs and obscures crucial information. While prompt compression methods have been proposed to alleviate these issues, they still suffer from the difficulty in capturing global context due to the lengthy input and training of a compressor. To overcome these challenges, we introduce a novel prompt compression method, \emph{Reading To Compression (R2C)}, which adopts the Fusion-in-Decoder architecture. Specifically, cross-attention scores are employed to identify essential tokens, effectively capturing global context and reducing computational overhead without compromising accuracy. Additionally, R2C naturally discerns important information via question-answer training, circumventing the necessity for pseudo-labels. Experimental results demonstrate that the proposed method significantly enhances efficiency without losing performance by effectively retaining critical information.
\end{abstract}

\iffalse
% v1
Large language models (LLMs) have demonstrated significant performance gains across various tasks due to advanced prompting techniques, such as Retrieval-Augmented Generation and Chain-of-Thought. 
However, the increasing length of prompts to convey sufficient information poses challenges, including higher computational costs and difficulty in discerning crucial information. 
Recently, prompt compression methods have been proposed to address these issues, but they suffer from limitations such as the difficulty of considering global context due to input length and a lack of optimal training data.
To overcome these challenges, we introduce a novel prompt compression method, Reading To Compression (R2C), which utilizes the Fusion-in-Decoder (FiD) architecture.
R2C employs cross-attention scores in the FiD decoder to identify and preserve essential tokens within lengthy inputs, effectively capturing global context and reducing computational overhead without compromising accuracy.
Additionally, R2C uses question-answer training to naturally capture important information, circumventing the necessity for unreliable pseudo-labels.
Experimental results on in-domain and out-of-domain datasets demonstrate that R2C significantly enhances LLM performance, reducing input length while retaining critical information.
\fi

%% file: sec-introduction.tex
\section{Introduction}\label{sec:introduction}

% -- 전체 구성 --
% 1. LLM 과 Lengthy input이 생길 수 있는 경우들 소개: CoT, ICL, RAG, ...
% 2. Prompt compression: soft prompting & token pruning
% 3. Extractive token pruning: i) entropy-based and ii) score-based approaches
% 4. lengthy input을 다루는 기존 Prompt Compression 모델의 챌린지: 
    % 4.1. How to extract essential information based on the global context
    % 4.2. How to train a compressor without optimal training data
% ------
% 5. 두 가지 챌린지를 해결하는 방안 소개: 
    % 5.1. FiD가 무엇이고 어떤 성질(multiple input aggregation)이 있는지
    % 5.2. FiD를 통한 QA 학습이 prompt compression과 어떻게 이어질 수 있는지 
% (question in QA == instruction   in ICL) 
% (context  in QA == demonstration in ICL) 
% QA에서는 question을 위한 정보를 context에서 찾아야하며, ICL에서는 instruction(end task)을 위한 정보를 demo에서 잘 찾아야 함.  
% 6. 제안 모델 소개 (R2C)

Large language models (LLMs) have recently exhibited remarkable performance gains in various tasks owing to a wide variety of prompting, \eg, Retrieval-Augmented Generation (RAG)~\cite{corr/abs-2312-10997/RAG-survey}, Chain-of-Thought (CoT)~\cite{nips/Wei0SBIXCLZ22/CoT}, and In-Context Learning (ICL)~\cite{abs-2301-00234/ICLsurvey}. The use of rich prompts unlocks the abilities of LLMs, but prompts can be verbose to deliver sufficient information. The lengthy prompts not only increase computational costs but also make LLMs struggle to discern important information. Although the input length limit has recently been extended to a million tokens~\cite{corr/abs-2403-05530/Gemini1.5}, the quadratic increase in computation over the input length is still a substantial burden.

\input{Figures_tex/fig_R2C_motivation}

Recently, \textit{prompt compression} has garnered significant focus for alleviating this issue. Its goal is to preserve only the essential information to reduce the computational overhead without sacrificing the accuracy of the end task. The most straightforward approach is \textit{token pruning}, removing redundant tokens from the original prompt by entropy-based metrics~\cite{emnlp/LiDGL23/Selective-Context, emnlp/JiangWLYQ23/LLMLingua, corr/abs-2310-06839/LongLLMLingua} or predicted scores~\cite{iclr/XuSC24/RECOMP, corr/abs-2312-08901/CoT-Influx, corr/abs-2403-12968/LLMLingua-2}. They utilize the perplexity of tokens using smaller models or classification scores of trained compressors. They can be used in a model-agnostic manner, delivering the benefit in black-box scenarios without understanding the internal structure of the LLMs.

However, existing prompt compression methods face two challenges in handling lengthy inputs.

\noindent
(i) \emph{How to extract essential information based on the global context?} 
The key points in a single paragraph may differ significantly from the main topic of the lengthy document. However, existing works divide the prompt into multiple segments if exceeding the maximum input length of compressors and compress each segment independently~\cite{corr/abs-2403-12968/LLMLingua-2, corr/abs-2312-08901/CoT-Influx}. Since the model identifies crucial tokens only within the segment, it is limited in capturing essential information across the global context.

\noindent
(ii) \emph{How to train a compressor?} 
Since the ground truth of the compressed prompt is non-trivial to define, it is difficult to train the compressor. Recent work detours the issue by utilizing state-of-the-art LLMs, \eg, GPT-4, for generating the pseudo-compressed prompts to train the compressor~\cite{corr/abs-2403-12968/LLMLingua-2}. However, as pointed out in \citet{emnlp/JiangWLYQ23/LLMLingua, corr/abs-2404-00489/Prompt-SAW}, GPT-4 underperforms in prompt compression, suggesting that there is still room for improvement in training compressors.

Addressing these issues, we shed light on one prominent solution to handle multiple documents, \ie, \emph{Fusion-in-Decoder (FiD)}~\cite{eacl/IzacardG21/FiD}. FiD is a question-answering model aggregating evidence from multiple documents to answer a question. It effectively captures the global context over multiple documents by leveraging cross-attention to answer the question, as depicted in Figure~\ref{fig:fig_R2C_motivation}. It is worth noting that FiD effectively highlights salient parts based on the global context to generate the answer regardless of the length of the whole context.

To this end, we propose a novel prompt compression method, \textit{\textbf{R}eading \textbf{To} \textbf{C}ompressing (\textbf{R2C})}, which fully utilizes the structure and training strategy of FiD to align with prompt compression. First, the prompt compression is connected with the FiD to capture the global context of lengthy input. Specifically, lengthy prompts are divided into multiple chunks as input units of FiD, and global semantics over chunks are aggregated in the decoder. R2C yields efficiency by utilizing the cross-attention scores computed in generating only the first token, avoiding auto-regressive generation. Second, we utilize the \textit{question-answering} task to train the compressor, a natural way to identify key information without relying on pseudo-labels. The cross-attention scores are trained to align with a question-answering process, providing the effective approximation of tokens that the target LLM needs to focus on.

We thoroughly conduct experiments on in-domain, \ie, Natural Questions, and out-of-domain datasets, \ie, LongBench, demonstrating the effectiveness and efficiency of R2C. Notably, R2C yields up to 14.5 times faster compression than existing methods and reduces end-to-end latency by 26\% with a 5x compression of the original prompt with minimal performance degradation.

%% file: Figures_tex/fig_R2C_motivation.tex
\begin{figure}[t]
\includegraphics[width=0.95\linewidth]{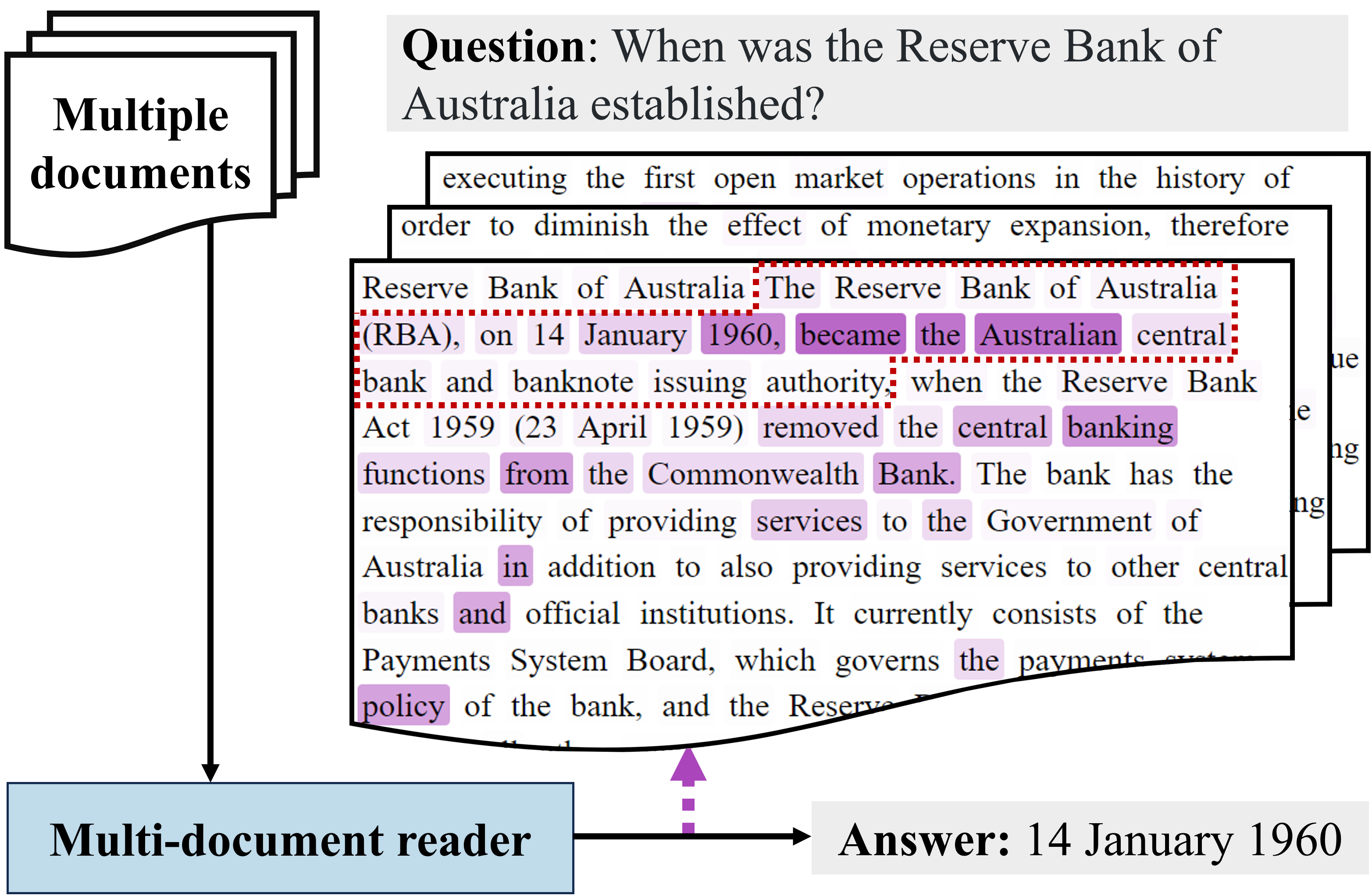}
\caption{Multi-document reader, \ie, Fusion-in-decoder (FiD), captures core information by learning to generate answers from lengthy inputs, as highlighted with the dotted red box. The darker the purple color, the higher the cross-attention score in FiD-decoder.}\label{fig:fig_R2C_motivation}
\vskip -0.2in
\end{figure}
% ~\cite{eacl/IzacardG21/FiD}

%% file: sec-related_work.tex
\section{Related Work} \label{sec:related_work}

\subsection{Prompt Compression}\label{sec:Prompt Compression}
As more complex prompting techniques are proposed to unlock the capabilities of LLMs, \textit{prompt compression} has been actively studied to handle lengthy input. Prompt compression methods are broadly categorized into soft prompting and two token pruning methods which are abstractive and extractive compression. 

\noindent
\textbf{Soft Prompting}. Soft prompting methods compress texts in embedding space, using a limited number of learnable embeddings to imply input tokens~\cite{emnlp/ChevalierWAC23/AutoCompressor, nips/Mu0G23/Gist, icml/QinD23/Nugget, corr/abs-2405-13792/xRAG}. They achieve high compression ratios with minimal loss of semantics. However, the embedding must be learned for each language model, and it is challenging to apply to API-based LLMs.

\noindent
\textbf{Abstractive Compression}. Abstractive compression aims to generate the core information of a prompt using a generative model. \citet{corr/abs-2311-08377/FilCO} introduced sentence-level pseudo-labels and trained the model to generate labeled sentences given the original prompt. \citet{corr/abs-2402-18700/Nano-Capsulator} proposed to optimize the compressor using a reward function considering length constraints. Recently, \citet{corr/abs-2404-00489/Prompt-SAW} proposed to construct a graph with LLMs and reconstruct the prompt using subgraphs within the graph. While abstractive compression effectively compresses prompts by reconstructing them, they suffer from the substantial cost of auto-regressive generation.

\noindent
\textbf{Extractive Compression}. Extractive compression methods extract only core information from prompts. Representative works proposed by \citet{emnlp/LiDGL23/Selective-Context, emnlp/JiangWLYQ23/LLMLingua} remove redundancy based on the entropy-based metric without any training. \citet{corr/abs-2310-06839/LongLLMLingua} additionally leverages question-aware perplexity. However, the entropy-based methods are hardly aligned with the objective of prompt compression, which is to retain only the essential information~\cite{corr/abs-2404-00489/Prompt-SAW}. Recently, there have been works on training compressors to extract the salient information from the prompt. \citet{corr/abs-2403-12968/LLMLingua-2, iclr/XuSC24/RECOMP} created pseudo labels and \citet{corr/abs-2312-08901/CoT-Influx, corr/abs-2308-08758/PCRL} incorporated reinforcement learning for training a compressor. However, they still struggle to capture the global context across whole prompts since each segment of prompts is compressed independently.

\subsection{Fusion-in-Decoder (FiD)}\label{sec:relatedwork_fid}
\noindent
Fusion-in-Decoder~\cite{eacl/IzacardG21/FiD} has been introduced for Open-Domain Question Answering (ODQA), effectively aggregating information from multi-documents to generate an answer. Recent studies have demonstrated the versatility of the FiD structure in various tasks thanks to its ability to handle lengthy input without information loss. \citet{iclr/IzacardG21/FiD-KD} showed that the cross-attention score obtained from the FiD decoder can be utilized as a label for retrieval. In addition, \citet{acl/YeBPRH23/FiD-ICL} incorporated the FiD structure for in-context learning with long input.
This paper introduces a new method for compressing long prompts utilizing FiD and demonstrates its effectiveness.

%% file: sec-method.tex
\section{Proposed Method} \label{sec:method}

This section introduces R2C, a novel prompt compression method using the Fusion-in-Decoder (FiD) architecture. In contrast to the existing compression methods, which consider the local context within each chunk, R2C effectively seizes the global context that lies across chunks in the lengthy prompt.

The prompt $\text{P}_\text{C}$ consists of an instruction $\text{I}$, a context $\text{C}$, and a question $\text{Q}$, \ie, $\text{P}_\text{C} = \text{(I; C; Q)}$, where the context is usually the longest component. R2C compresses the context $\text{C}$ to $\hat{\text{C}}$ and reduces the overall prompt length from $|\text{P}_\text{C}|$ to $|\text{P}_{\hat{\text{C}}}|$.

Figure~\ref{fig:fig_R2C_framework} depicts the overall framework of R2C. First, we divide context $\text{C}$ into multiple chunks and feed them into FiD to obtain the importance score for each token (Section~\ref{sec:FiD}). Next, token scores are aggregated in chunk- and sentence-level for multi-granularity compression (Section~\ref{sec:unit_importance}). Finally, we hierarchically compress the context in a coarse-to-fine manner, \ie, chunk-to-sentence order (Section~\ref{sec:hierarchical compression}). Note that the training process is omitted in this paper since R2C is not trained for prompt compression. Instead, R2C utilizes the trained FiD weights for the QA task.

\input{Figures_tex/fig_R2C_framework}

\subsection{Identifying Importance in Context}\label{sec:FiD}

We calculate the token-level importance in the context using FiD~\cite{eacl/IzacardG21/FiD}. The context is divided into smaller chunks as input units, and each chunk is processed individually with multiple encoders. The outputs of multiple encoders are concatenated and utilized as a key-value matrix of cross-attention for decoding. Even if the overall length of a prompt exceeds the maximum input length of language models, FiD discerns the important parts considering the global context. Here, we leverage the attention score as the importance criterion.

Given a prompt $\text{P}_\text{C}$, we divide the context $\text{C}$ into $K$ chunks: $\mathcal{C}$ = $[\text{C}_1, \text{C}_2, ..., \text{C}_K]$.
For each chunk $\text{C}_i$, we input it to the FiD-encoder and get the token embeddings $\mathbf{H}_i\in \mathbb{R}^{M \times h}$, where $M$ is the maximum sequence length of the FiD-encoder, and $h$ is the size of the hidden dimension. 
Similar to the original QA task, we prepend the question $\text{Q}$ to $\text{C}_i$ if a question is given in the dataset, otherwise, we set $\text{Q}$ as an empty string. 
    \begin{equation}
    \mathbf{H}_i=\text{FiD-encoder}\left(\text{Q} + \text{C}_i\right) \in \mathbb{R}^{M \times h}. \\
    \end{equation}

We perform it for $K$ chunks and concatenate all token embeddings as $\mathbf{H} = (\mathbf{H}_1; \dots; \mathbf{H}_K) \in \mathbb{R}^{(K \times M) \times h}$.
Then, $\mathbf{H}$ is converted to the key matrix $\mathbf{K}$ through the projection layer $\mathbf{W}_k \in \mathbb{R}^{h \times h}$.
The cross-attention score $\mathbf{A} \in \mathbb{R}^{1 \times (K \times M)}$ is calculated by the matrix product using query embedding $\mathbf{q}_{\texttt{[BOS]}} \in \mathbb{R}^{1 \times h}$ and the key matrix $\mathbf{K}$.
    \begin{equation}
    \mathbf{A} = \text{softmax} \left (\frac{\mathbf{q}_{\texttt{[BOS]}} \mathbf{K}}{\sqrt{h}} \right ), \text{where} \  \mathbf{K} = \mathbf{W}_k \mathbf{H}^\top.
    \end{equation}

Due to the maximum sequence length, existing works~\cite{emnlp/JiangWLYQ23/LLMLingua, corr/abs-2310-06839/LongLLMLingua} are limited to the local context within chunks. In contrast, R2C effectively captures the global context across all chunks by processing concatenated outputs. We define the token-level importance ${t}_{i, j}$ of the $j$-th token in $i$-th chunk as the sum of the attention scores $\mathbf{A}$ over all layers and heads in the FiD-decoder.
    \begin{equation}
    t_{i, j} = \sum_{l=1}^{L} \sum_{h=1}^{H} \textbf{A}_{i,j}^{(l,h)}.
    \end{equation}
Here, $L$ is the number of layers in the decoder, and $H$ is the number of heads.
$\textbf{A}_{i,j}^{(l,h)}$ denotes the attention score for the $j$-th token in the $i$-th chunk of $h$-th head in $l$-th layer.

\subsection{Aggregating Unit Importance}\label{sec:unit_importance} 
We adopt two compression units with coarser granularity than tokens, \ie, chunks and sentences. A naive way of compression is to prune redundant tokens with low importance scores until the desired compression ratio is reached. However, token-level compression neglects the semantic integrity and the grammatical structure of the text~\cite{emnlp/JiangWLYQ23/LLMLingua, corr/abs-2404-00489/Prompt-SAW}. Otherwise, R2C adopts two levels of granularity, achieving high compression ratios without hindering the semantics.

\noindent
\textbf{Chunk-level Importance.}
We utilize an average pooling to aggregate chunk-level importance following \citet{iclr/IzacardG21/FiD-KD}. Note that other pooling operators, \eg, max or sum, are also feasible and please refer to Table~\ref{tab:ablation} for further experimental analysis. The chunk-level importance $c_i$ is averaged over the token-level scores $t_{i, j}$ contained in each chunk $\text{C}_i$.
    \begin{equation}
    c_i = \frac{1}{|\text{C}_i|}\sum_{j=1}^{|\text{C}_i|} t_{i, j}, 
    \end{equation}
where $|\text{C}_i|$ denotes the number of tokens in chunk $\text{C}_i$. 
We sort the chunks $\text{C}$ in descending order of chunk-level scores and re-index them.

\noindent
\textbf{Sentence-level Importance.} 
For a fine-grained unit, R2C also utilizes sentence-level importance. A sentence is a basic unit that preserves the meaning of the original input while compression~\cite{corr/abs-2311-08377/FilCO, iclr/XuSC24/RECOMP}.

We split each chunk $\text{C}_i$ to the sentence list $\mathcal{S}_i$$=[\text{S}_{i,1}, \text{S}_{i,2}, \dots, \text{S}_{i,|\mathcal{S}_i|}]$ using \texttt{sent\_tokenize} of NLTK~\cite{lre/Wagner10/NLTK}.
For the $m$-th sentence in $i$-th chunk $\text{S}_{i, m}$, we compute the sentence-level importance $s_{i, m}$ as an average over the scores of contained tokens.
    \begin{equation}
    \begin{gathered}
    s_{i, m} = \frac{1}{|\text{S}_{i, m}|}\sum_{j=1}^{|\text{C}_{i}|} t^{m}_{i, j}, \\ 
    \text{where} \ \ t^{m}_{i, j} = 
        \begin{cases}
            t_{i,j} & \text{if $j$-th token $\in$ S$_{i, m}$}, \\
            0       & \text{otherwise}. \\
        \end{cases} \\
    \end{gathered}
    \end{equation}
Here, $|\text{S}^{m}_{i}|$ denotes the number of tokens in the sentence $\text{S}^{m}_{i}$.
Then, similar to chunk-level, we sort and re-index the sentences in $\mathcal{S}_i$ according to sentence-level importance scores.

\noindent
\textbf{Token-level Importance.} It is possible to utilize token-level importance as the finest granularity. Although we derive the token-level scores in R2C, it is observed that token-level compression may hinder the performance of the target task. Therefore, R2C focuses on utilizing chunk- and sentence-level importance without losing semantic integrity.

\subsection{Performing Hierarchical Compression}\label{sec:hierarchical compression} 
Given the multi-granularity unit importance, \ie, chunk and sentence, R2C hierarchically compresses the prompt. To preserve semantic integrity~\cite{corr/abs-2310-06839/LongLLMLingua, corr/abs-2312-08901/CoT-Influx}, it performs compression from chunks to sentence excluding token-level. Token-level compression can break the grammatical structure or struggle to generate answers exactly matched to the ground truth\footnote{With token-level compression, "\textit{She moved to Los Angeles, where she studied drama at the Lee Strasberg Theatre and Film Institute}" becomes "\textit{Los Angeles studied drama at the Lee Strasberg Theatre and Film Institute}", destroying the linguistic structure and semantics.}.

Algorithm \ref{alg:coarse} describes the compression procedure of R2C.
Given sorted chunks $\mathcal{C}$ as input, R2C first performs chunk-level compression and generates $\hat{\mathcal{C}}_{\text{chunk}}$ by retaining only the important chunks (line \ref{alg_line:chunk_start}--\ref{alg_line:chunk_end}).
Then, it further performs sentence-level compression and selects only the crucial sentences for each chunk to yield $\hat{\mathcal{C}}_{\text{sent}}$ (line \ref{alg_line:sent_start}--\ref{alg_line:sent_end}). We use $\hat{\mathcal{C}}_{\text{sent}}$ for the final compressed context $\hat{\text{C}}$ (line~\ref{alg_line:final_concat}), otherwise $\hat{\mathcal{C}}_{\text{chunk}}$ is used if sentence-level compression is not applied. When concatenating $\hat{\mathcal{S}}_{i}$ to $\hat{\mathcal{C}}_{sent}$ (line~\ref{alg_line:sent_concat}), we restore them to their original order, except for $\hat{\mathcal{C}}_{sent}$ in Natural Questions due to lost-in-the-middle problem~\cite{corr/abs-2307-03172/lost-in-the-middle}.

\input{Figures_tex/algorithm1_coarse}

\noindent
\textbf{Step 1. Chunk-level Compression.}
We first determine the hierarchical ratio between two-level compression using a hyperparameter $\rho$.
Given the length of the original prompt be $|\text{P}_\text{C}|$ and the number of target tokens $T$, the number of removing tokens is defined as $E_{\text{comp}}=|\text{P}_\text{C}|-T$.
We set the number of chunk- and sentence-level compression tokens as $E_{\text{chunk}}=\rho \cdot E_{\text{comp}}$ and $E_{\text{sent}}=(1-\rho) \cdot E_{\text{comp}}$, respectively.
That is, if $\rho$ is 1, R2C only performs chunk-level compression, and if $\rho$ is 0, R2C only performs sentence-level compression.
R2C constructs a compressed chunk list $\hat{\mathcal{C}}_{\text{chunk}}$ by adding chunks until the number of removing tokens $E_{\text{chunk}}$ is exceeded by the remaining tokens.

\noindent
\textbf{Step 2. Sentence-level Compression.}
R2C adaptively sets the number of compressed sentences for each chunk based on chunk-level importance. It allows R2C to reserve more information in more salient chunks, reflecting the global context. 
    \begin{equation}\label{eq_e_sent}
    \begin{gathered}
    E_{\text{sent}, i} =  \frac{(1 / \hat{c}_{i})^{\gamma}}{\sum_{\hat{c}_{i} \in \hat{c}} (1 / \hat{c}_{i})^{\gamma}} \times {E_{\text{sent}}}, \\
    \end{gathered}
    \end{equation}
$\gamma$ is a coefficient that controls the impact of inverted chunk-level scores. Note that if $\gamma$ is 0, the sentence compression is applied uniformly to all chunks.

%% file: Figures_tex/fig_R2C_framework.tex
\begin{figure*}[t]
\includegraphics[width=1.0\linewidth]{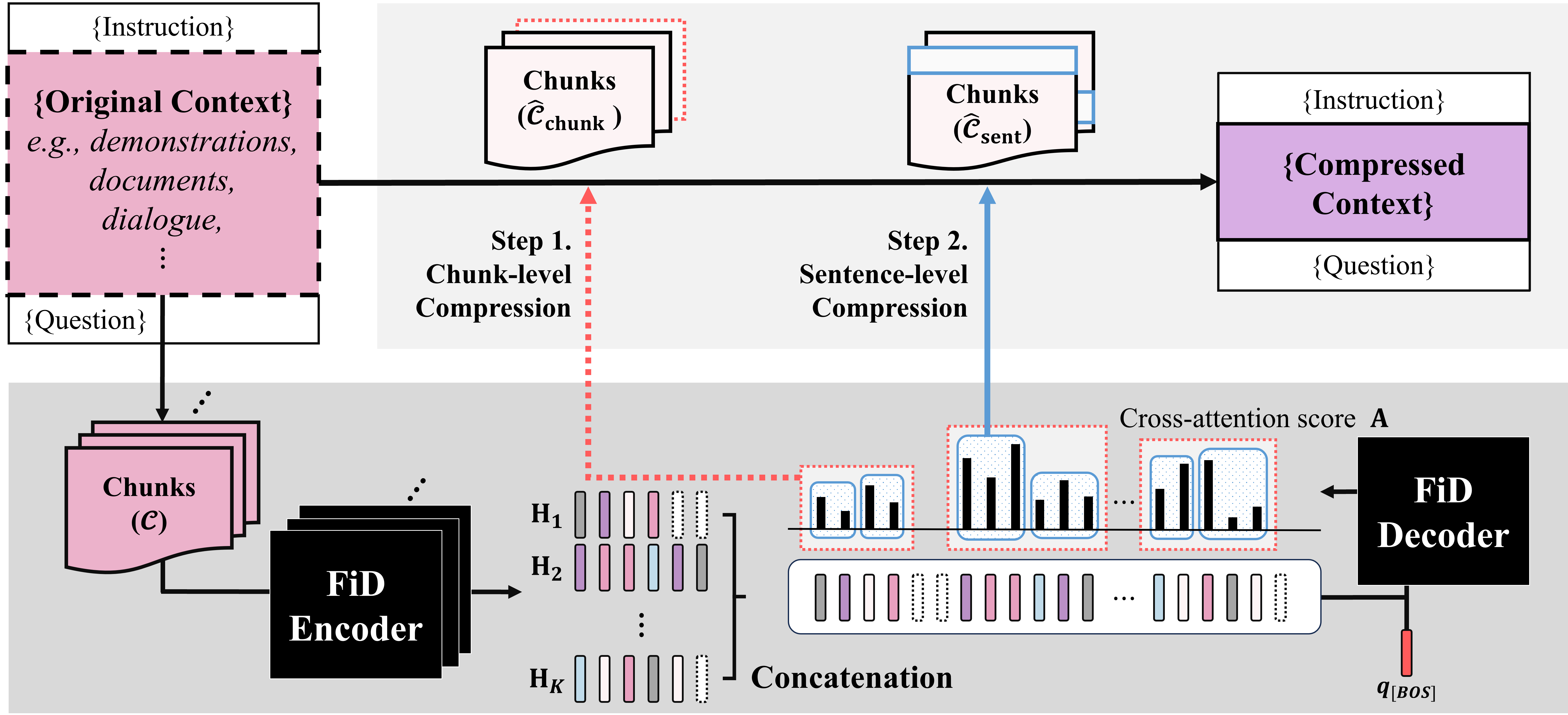}
% \vskip 0.5in
\caption{The overall framework of \textbf{R}eading \textbf{To} \textbf{C}ompressing (\textbf{R2C})}\label{fig:fig_R2C_framework}
\vskip -0.1in
\end{figure*}

%% file: Figures_tex/algorithm1_coarse.tex
\begin{algorithm}[t]
\caption{R2C prompt compression}\label{alg:coarse}
\small
\begin{algorithmic}[1]
\Require 
sorted context chunks $\mathcal{C}=[\text{C}_{1}, \text{C}_{2}, ... \text{C}_{K}]$, number of chunks $K$, number of total removing tokens $E_{\text{comp}}$, hierarchical ratio between two-level compression $\rho$
\Ensure compressed context $\hat{\text{C}}$

\CommentAligned{▶ Step 1. Chunk-level Compression}
\State $E_{\text{chunk}}=\rho \cdot E_{\text{comp}}$ \label{alg_line:chunk_start}
\For {$i = 1$ \textbf{to} $K$}
    \If {$ E_{\text{chunk}} \geq \sum_{j=i}^{K} |\text{C}_j| $}
        \State \textbf{Break}
    \EndIf
    \State $K'=K'+1$
\EndFor
\State $\hat{\mathcal{C}}_{\text{chunk}} \gets [\text{C}_{1}, \text{C}_{2}, ... \text{C}_{K'}]$ \label{alg_line:chunk_end}

\CommentAligned{▶ Step 2. Sentence-level Compression}
\State Initialize $\hat{\mathcal{C}}_{\text{sent}} \gets [~]$ \label{alg_line:sent_start}
\State $E_{\text{sent}}=(1-\rho) \cdot E_{\text{comp}}$
\For {$i = 1$ \textbf{to} $K'$}
    \State Calculate $E_{\text{sent}, i}$ using Equation~\eqref{eq_e_sent}
    \State ${\mathcal{S}}_{i} = {\texttt{sent\_tokenize}}({\text{C}}_{i})$ 
    \State Initialize $\hat{\mathcal{S}_i} \gets [~]$
    \For {$m = 1$ to $|{\mathcal{S}}_{i}|$}
        \If {$ E_{\text{sent}, i} \geq \sum_{j=m}^{|{\mathcal{S}}_{i}|} |{\text{S}}_{i, j}| $}
            \State \textbf{Break}
        \EndIf
        \State Add ${\text{S}}_{i,m}$ to $\hat{\mathcal{S}_i}$
    \EndFor
    \State $\hat{\text{C}}_{i} = \texttt{concatenate}(\hat{\mathcal{S}_i})$ \label{alg_line:sent_concat}
    \State Add $\hat{\text{C}}_{i}$ to $\hat{\mathcal{C}}_{\text{sent}}$
\EndFor \label{alg_line:sent_end}
\State $\hat{\text{C}} = \texttt{concatenate}(\hat{\mathcal{C}}_{\text{sent}})$ \label{alg_line:final_concat}

\end{algorithmic}
\end{algorithm}

%% file: sec-exp_setup.tex
\section{Experiments Setup}\label{sec:exp_setup}
\subsection{Datasets}
We validate the performance of R2C on two datasets.
(i)  \textbf{In-domain}: We utilize Natural Questions (NQ)~\cite{tacl/KwiatkowskiPRCP19/NQ}, which are widely adopted in Open-domain Question Answering (ODQA) tasks. We retrieve 20 candidate passages for each question using DPR~\cite{emnlp/KarpukhinOMLWEC20/DPR, iclr/IzacardG21/FiD-KD}. 
(ii) \textbf{Out-of-domain}: To evaluate the generalizability of compressed prompts, we use the LongBench~\cite{bai2023longbench}\footnote{\url{https://github.com/THUDM/LongBench}}. We include five types of tasks: single-document QA (SingleDoc), multi-document QA (MultiDoc), summarization (Summ.), few-shot learning (FewShot), and code completion (Code). Note that we only evaluate the English datasets and omit the synthetic tasks to validate the model's ability in real-world scenarios. (See appendix~\ref{sec:app_dataset} for further details.)

\subsection{Evaluation Metrics and Prompts}
For Natural Questions, we use Span Exact Match (Span EM) and prompts following \citet{corr/abs-2307-03172/lost-in-the-middle} to evaluate whether the generated text contains the answer. For LongBench, we follow metrics and prompts of each dataset provided by the official benchmark (Refer to appendix~\ref{sec:app_metric} and~\ref{sec:app_prompt}).

\subsection{Baselines}
We compared R2C with the following models. (i) Two retrieval-based models: BM25~\cite{SIGIR/RobertsonW94/BM25} and DPR~\cite{emnlp/KarpukhinOMLWEC20/DPR}, performing only chunk-level compression. DPR~\cite{emnlp/KarpukhinOMLWEC20/DPR} is trained with knowledge distillation on the NQ dataset~\cite{iclr/IzacardG21/FiD-KD}. For the BM25 results in the LongBench dataset, we follow the experimental setup from LongLLMLingua~\cite{corr/abs-2310-06839/LongLLMLingua}. The key difference is that we apply BM25 at the chunk-level instead of the sentence-level. (ii) Five compression-based models: Selective-Context~\cite{emnlp/LiDGL23/Selective-Context}, LLMLingua~\cite{emnlp/JiangWLYQ23/LLMLingua}, LongLLMLingua~\cite{corr/abs-2310-06839/LongLLMLingua}, LLMLingua-2~\cite{corr/abs-2403-12968/LLMLingua-2} and RECOMP~\cite{iclr/XuSC24/RECOMP}. Unlike other baseline methods, RECOMP compresses the context at the sentence-level by selecting sentences based on the similarity between the question and sentence embeddings. As a result, RECOMP is unsuitable for tasks where a question is absent, such as summarization, and is reported only on QA tasks.

\subsection{Implementation Details}
We trained the Fusion-in-Decoder model~\cite{eacl/IzacardG21/FiD} on the Natural Questions train set, utilizing 20 passages for each question. For target LLMs, we used Llama2-7b-chat-hf~\footnote{\url{https://huggingface.co/meta-llama}} (LLaMA2-7B, \citealp{journals/corr/abs-2307-09288/llama2}) and GPT-3.5-turbo-1106~\footnote{\url{https://chatgpt.com/}} (GPT-3.5). We randomly sampled 20\% of the NQ dev set to tune the compression hyper-parameters in R2C. We set the hierarchical ratio $\rho$ and the chunk-level score coefficient $\gamma$ as 0.8 and 1, respectively. All token counts are based on the ChatGPT tokenizer. We regard a demonstration, paragraph, or dialogue as a chunk unit if given. For longer chunks, we split them based on line breaks and set the maximum token length for each chunk as 128. We reproduced baselines based on the official codes and reported with the results from the original papers. 

%% file: sec-results.tex
\section{Results and Analysis} \label{sec:results}

\subsection{Main Results}

\noindent\textbf{Question Answering Task}. 
Table~\ref{tab:nq} reports the accuracy of FiD and two target LLMs, namely GPT-3.5 and LLaMA2-7B, with different compression methods on the Natural Questions test set. (i) The proposed method achieves the best performance compared to state-of-the-art compression methods. Specifically, it shows improvements of 5.6\% and 11.1\% over the most effective baseline, RECOMP~\cite{iclr/XuSC24/RECOMP}, when GPT-3.5 and LLaMA2-7B are used as target LLMs, respectively. Notably, R2C outperforms the original prompt when used with LLaMA2, and with GPT-3.5, we achieve comparable performance with six times fewer tokens. (ii) DPR~\cite{emnlp/KarpukhinOMLWEC20/DPR}, RECOMP~\cite{iclr/XuSC24/RECOMP}, and R2C, which are trained on question-answer datasets, \ie, in-domain evaluation, significantly outperform other methods. While the other two methods are optimized to identify significant passages or sentences, R2C yields better performance. This suggests that learning to answer questions directly contributes to capturing important information in context, and highlights the benefits of using cross-attention scores for compression after QA training. (iii) Despite the relatively modest size of FiD (223M) in comparison to LLaMA-2 (7B), it performs well on the QA task, with only a 4.5\% difference in accuracy. This indirectly suggests that FiD is effective at capturing important information within multiple documents.

\input{Tables/tab_nq}

\input{Tables/tab_longbench}

\noindent\textbf{Out-of-domain Evaluation}. 
Table~\ref{tab:longbench} shows an accuracy of R2C and other compression methods with two target LLMs on out-of-domain datasets in LongBench~\cite{bai2023longbench}\footnote{While the official results without compression (\ie, Original and Original*) are presented, they are not fully evaluated. This is because the middle part has been truncated if samples exceed the maximum sequence length.}. Note that RECOMP selects sentences based on their similarity to the question embedding, so it was only evaluated on the QA task. We summarize our key observations as follows:

(i) R2C yields the highest average performance for both target LLMs with improvements of 3.0\% in GPT-3.5 and 9.2\% in LLaMA2-7B over the best competitive baseline. It also outperforms the original prompt in two QA tasks, SingleDoc and MultiDoc, indicating the effectiveness of removing the ambiguity of lengthy prompts using R2C.
(ii) The question-answering task proves to be an effective alternative for compressor training. R2C shows superior performance on all tasks, including QA tasks. We also report the performance of each model as reported in their respective papers. Inconsistencies in the results are likely due to differences in the version of GPT-3.5. Focusing on the results without an asterisk (-), which we reproduced using the same version, GPT-3.5-turbo-1106, R2C achieved the highest performance among compression-based methods. Additionally, we observed that R2C performed well on the code completion task, likely because generating code from multiple files is analogous to generating answers from multiple documents.
(iii) We observe different trends depending on the target LLM for the FewShot task. Specifically, on GPT-3.5, chunk-level filtering with BM25 outperforms R2C, achieving 68.1 compared to 66.9. However, on LLaMA2-7b, R2C significantly surpasses BM25, with scores of 64.5 and 61.5, respectively. It suggests that GPT-3.5 can understand the task even when the demonstration is not directly relevant, highlighting the importance of using the intact demonstration. R2C, by compressing hierarchically, may disrupt the demonstration. We conjecture that different prompts may require different levels of granularity for effective compression.

\subsection{In-depth Analysis}
\noindent\textbf{Compression Efficiency}.
Figure~\ref{fig:fig_latency_targetlength}-(a) presents the efficiency of R2C and baseline methods on LongBench. We measure the latency using a single NVIDIA A6000 GPU with a batch size of 1. The y-axis represents the average accuracy of the tasks included in LongBench, and the x-axis represents the average compression latency per prompt. Results are based on $T=2000$ and GPT-3.5 as the target LLM. Notably, R2C dramatically improves efficiency while also enhancing accuracy, achieving a latency improvement of 1.6 times over the most efficient existing method, \ie, LLMLingua-2, and 14.5 times over the most effective method, \ie, LongLLMLingua. This improvement in efficiency is largely due to the smaller model size used for compression. Specifically, R2C is based on T5-base~\cite{JMLR/Raffel2020/T5}, which has 223M parameters, while LLMLingua-2 adopts XLM-RoBERTa-large~\cite{acl/ConneauKGCWGGOZ20/XLMRobertalarge} with 355M parameters. LongLLMLingua and LLMLingua use a large LLM~\cite{journals/corr/abs-2307-09288/llama2} with 7B parameters, resulting in significant computational overhead. Additionally, although R2C uses a generative model, it efficiently captures important information by utilizing only the cross-attention scores from the first token generation. The results demonstrate that R2C enhances the quality of compressed prompts and optimizes the compression efficiency.

\input{Figures_tex/fig_latency_targetlength}
\input{Tables/tab_e2e_efficiency}

\noindent\textbf{End-to-end Efficiency}. 
Table~\ref{tab:e2e_efficiency} illustrates the end-to-end latency of R2C depending on the different number of target tokens. We use 3,350 samples in LongBench datasets and set the maximum decoding token of the API to 200 for all datasets following~\cite{corr/abs-2310-06839/LongLLMLingua}. Although token pruning can be used in API-based models and reduce API costs, the benefits can be offset if the compression process takes too long. Notably, the overall end-to-end inference time is accelerated with the proposed method. As shown in Figure~\ref{fig:fig_latency_targetlength}-(a), the compression is efficiently performed by R2C. Considering the latency of both compression and the API, R2C can infer the answer with 74.0\% latency compared to the original prompt.

\noindent\textbf{Varying Target Length}. Figure~\ref{fig:fig_latency_targetlength}-(b) shows the average performance on LongBench according to the length of compressed prompts. We adjusted the compression ratio from 2x to 10x (\ie, 5K--1K tokens) for LongBench, which averages about 10K tokens originally. At all compression ratios, R2C delivers higher performance compared to the original prompt, indicating its ability to eliminate noisy information. However, LongLLMLingua exhibits lower performance at high compression ratios. Unlike R2C, LongLLMLingua employs entropy-based metrics, struggling with retaining important information at high compression ratios. Since the two QA tasks show different tendencies compared to the other tasks, we also report the full results for each task of R2C with different lengths in Table~\ref{tab:longbench_varying_t} in the appendix. The results show that the QA tasks, including SingleDoc and MultiDoc, achieve their best performance when the target length is in the range of 2000 to 3000 tokens, while the performance of the other tasks continues to improve as the target length increases. These results show that R2C not only outperforms other models due to its strong performance on QA tasks, but also excels at compression across a variety of tasks, resulting in superior generalization capabilities. Furthermore, this suggests that the performance of the target LLM can be further improved by first denoising the input contexts and then feeding them into the target model. 

\input{Tables/tab_ablation}
\subsection{Ablation Study}
\noindent\textbf{Comparison with variants of R2C}. Table~\ref{tab:ablation} analyzes the effectiveness of various strategies utilized in R2C. (i) R2C with token-level compression drastically degrades the performance of R2C. This indicates that since it focuses only on prominent tokens while neglecting semantic consistency, it easily confuses the target LLM, as shown in Section~\ref{sec:hierarchical compression}. (ii) R2C without the QA task training, \ie, T5-base~\cite{JMLR/Raffel2020/T5} initialization, shows a significant performance drop of 23.2\%. It implies that training the compressor with the QA task is appropriate as it naturally learns to discern salient parts in lengthy inputs. (iii) While the last decoder layer can focus on generating the final answer tokens, using cross-attention scores from all layers contributes to a performance gain. We attribute this to the tendency of the last decoder layer to focus excessively only on certain parts. (iv) Lastly, R2C aggregates unit importance by averaging token-level scores within a chunk or sentence. The results imply that average pooling is slightly more effective while using maximum token importance is valid. We adopt averaging as it captures crucial information across the entire unit and can provide a more balanced and comprehensive representation rather than focusing solely on the most prominent tokens.

\input{Figures_tex/fig_hyperparams}

\noindent\textbf{Effectiveness of Hierarchical Compression}.
Figure~\ref{fig:fig_hyperparams} illustrates the impact of hierarchical compression parameters (\ie, $\rho$ and $\gamma$) on the NQ dev set. 
Figure~\ref{fig:fig_hyperparams}-(a) illustrates that the balance of chunk and sentence-level compression by $\rho = 0.8$ yields the best performance, indicating that it can preserve essential information more effectively by considering both coarse- and fine-grained levels. Figure~\ref{fig:fig_hyperparams}-(b) depicts the effect of $\gamma$ in eq~\eqref{eq_e_sent}. Higher $\gamma$ removes more tokens from less important chunks. While $\gamma=0$ treats all chunks equally, using chunk-level importance in sentence-level compression improves performance by tailoring compression to the importance of each chunk. However, high $\gamma$ values reduce performance, suggesting that aggressive sentence-level compression on low-scored chunks harms overall compression quality.

%% file: Tables/tab_nq.tex
\begin{table}[t] \small
\centering
\begin{tabular}{c|c|c|c}
\toprule
\multicolumn{1}{c|}{\begin{tabular}[c]{@{}c@{}} Target \\ LLM \end{tabular}} & \multicolumn{1}{c|}{Compression}       & \multicolumn{1}{c|}{\begin{tabular}[c]{@{}c@{}}NQ test \\ (Span EM)\end{tabular}} & \# tokens \\ \midrule
FiD                                                          & -                   & 50.5                                                   & -         \\ \midrule
\multirow{9}{*}{GPT-3.5}                                              & Original                   & 66.7                                                   & 3,018     \\ \cmidrule{2-4} 
                                                                      & BM25                & 49.4                                                   & 534       \\
                                                                      & DPR                 & \uline{63.0}                                                   & 501       \\
                                                                      & Selective-Context   & 44.4                                                   & 501       \\
                                                                      & LLMLingua           & 41.9                                                   & 478       \\
                                                                      & LLMLingua-2         & 52.1                                                   & 510       \\
                                                                      & LongLLMLingua       & 55.6                                                   & 489       \\
                                                                      & RECOMP              & \uline{63.0}                                                   & 500       \\ \cmidrule{2-4} 
                                                                      & \textbf{R2C (ours)} & \textbf{66.5}                                          & 482       \\ \midrule
\multirow{9}{*}{\begin{tabular}[c]{@{}c@{}}LLaMA2\\ -7B\end{tabular}} & Original                   & 52.8                                                   & 3,018     \\ \cmidrule{2-4} 
                                                                      & BM25                & 41.8                                                   & 534       \\
                                                                      & DPR                 & \uline{54.3}                                                   & 501       \\
                                                                      & Selective-Context   & 38.1                                                   & 501       \\
                                                                      & LLMLingua           & 32.7                                                   & 478       \\
                                                                      & LLMLingua-2         & 42.5                                                   & 510       \\
                                                                      & LongLLMLingua       & 49.0                                                   & 489       \\
                                                                      & RECOMP              & 53.7                                                   & 500       \\ \cmidrule{2-4} 
                                                                      & \textbf{R2C (ours)} & \textbf{59.7}                                          & 482       \\ \bottomrule
\end{tabular}
\caption{Accuracy of FiD, GPT-3.5, and LLaMA2-7B in NQ test with 6x compressed prompt (\ie, $T=500$). For context $\text{C}$ of prompt P$_\text{C}$, we used 20 passages retrieved by DPR~\citeyearpar{emnlp/KarpukhinOMLWEC20/DPR, iclr/IzacardG21/FiD-KD}. The best and the second-best performance using the compressed prompt are marked in \textbf{bold} and \uline{underline}, respectively.}
\label{tab:nq}
\end{table} 

% iclr/IzacardG21/FiD-KD
% Model 대신 Retreiver | Reranker
% 왼쪽 정렬

% \begin{tabular}{c|cc}
% \toprule
% Model & R@1 & AUC \\ \midrule
% DPR & 49.9 & - \\ \midrule
% FiD (cross-attention) &  58.6 & - \\ \midrule
% MGFiD (w/ Pas\mathcal{L}_{\text{sent}}anker) & 61.7& - \\
% \multicolumn{1}{l|}{\hspace{6pt}$+\mathcal{L}_{\text{\mathcal{L}_{\text{sent}} \text{w/ Cross-entropy}$} & 60.9 & 0.70 \\
% \multicolumn{1}{l|}{\hspace{6pt}$+\mathcal{L}_{\text{sent}} \text{w/ Focal loss}$} & \textbf{62.2} & \textbf{0.82} \\
% \multicolumn{1}{l|}{\hspace{6pt}$+\mathcal{L}_{\text{sent.}} \text{w/ Cross-entropy}$} & 60.9 & 0.70 \\
% \multicolumn{1}{l|}{\hspace{6pt}$+\mathcal{L}_{\text{sent.}} \text{w/ Focal loss}$} & \textbf{62.2} & \textbf{0.82} \\

%% file: Tables/tab_longbench.tex
\begin{table*}[t]\small
\centering
\begin{tabular}{c|c|ccccc|c|c}
\toprule
Target LLM                 & Compression       & SingleDoc     & MultiDoc      & Summ.         & FewShot       & Code          & Average       & \# tokens \\ \midrule
\multirow{11}{*}{GPT-3.5}  & Original        & 43.2          & 46.1          & 25.2          & 69.2          & 64.4          & 49.6          & 9,881     \\
                           & Orignal*                & 39.7          & 38.7          & 26.5          & 67.0          & 54.2          & 45.2          & 9,881     \\ \cmidrule{2-9} 
                           & BM25              & 34.9          & 41.0          & 23.3          & \underline{68.1}    & 49.6          & 43.4          & 1,949     \\
                           & LLMLingua         & 30.4          & 31.4          & 20.9          & 66.0          & 55.0          & 40.7          & 1,830     \\
                           & Selective-Context & 29.8          & 35.4          & 22.1          & 52.4          & 45.0          & 36.9          & 2,009     \\
                           & LLMLingua-2       & 36.2          & 40.9          & 23.2          & 61.5          & 47.6          & 41.9          & 2,023     \\
                           & LLMLingua-2*      & 29.8          & 33.1          & \underline{25.3}    & 66.4          & \textbf{58.9} & 42.7          & {\tiny $\approx$}1,898    \\
                           & LongLLMLingua     & 37.0          & 44.9          & 22.0          & 65.1          & 49.4          & 43.7          & 1,743     \\
                           & LongLLMLingua*    & 39.0          & 42.2          & \textbf{27.4} & \textbf{69.3} & 56.6          & 46.9          & {\tiny $\approx$}1,753    \\
                           & RECOMP            & \underline{40.1}    & \underline{48.1}    & -             & -             & -             & -             & -         \\ \cmidrule{2-9} 
                           & \textbf{R2C (ours)}        & \textbf{43.5} & \textbf{48.7} & 24.9          & 66.9          & \underline{57.6}    & \textbf{48.3}          & 1,976     \\ \midrule
\multirow{9}{*}{LLaMA2-7b} & Original                 & 25.6          & 22.4          & 24.6          & 62.9          & 55.2          & 38.1          & 9,881     \\
                           & Original*                & 24.9          & 22.6          & 24.7          & 60.0          & 48.1          & 36.1          & 9,881     \\ \cmidrule{2-9} 
                           & BM25              & 25.4          & 25.4          & \underline{25.0}    & \underline{61.5}    & 44.0          & 36.3          & 1,949     \\
                           & LLMLingua         & 19.4          & 17.3          & 22.3          & 61.0          & \underline{51.0}    & 34.2          & 1,830     \\
                           & Selective-Context & 21.0          & 19.7          & 23.5          & 46.1          & 34.0          & 28.9          & 2,009     \\
                           & LLMLingua-2       & 22.6          & 23.8          & 23.8          & 56.0          & 43.2          & 33.9          & 2,023     \\
                           & LongLLMLingua     & 26.9          & 29.6          & 23.4          & 61.3          & 43.9          & \underline{37.0}    & 1,743     \\
                           & RECOMP            & \underline{27.4}    & \textbf{31.3} & -             & -             & -             & -             & -         \\ \cmidrule{2-9} 
                           & \textbf{R2C (ours)}        & \textbf{28.5} & \underline{29.8}    & \textbf{25.3} & \textbf{64.4} & \textbf{54.0} & \textbf{40.4} & 1,976     \\ \bottomrule
\end{tabular}
\caption{Performance of two LLMs (GPT-3.5 and LLaMA2-7B) in LongBench benchmark with 5x compressed prompt (\ie, $T=2,000$). \# tokens denotes the average number of tokens across all datasets based on the ChatGPT tokenizer. Among the compression methods, the best performance is marked in \textbf{bold} and the second-best is \underline{underlined}. * denotes the result from \cite{bai2023longbench, corr/abs-2310-06839/LongLLMLingua, corr/abs-2403-12968/LLMLingua-2}. }
\label{tab:longbench}
\end{table*}

%% file: Figures_tex/fig_latency_targetlength.tex
\begin{figure}[t]
\centering
\includegraphics[width=0.48\textwidth]{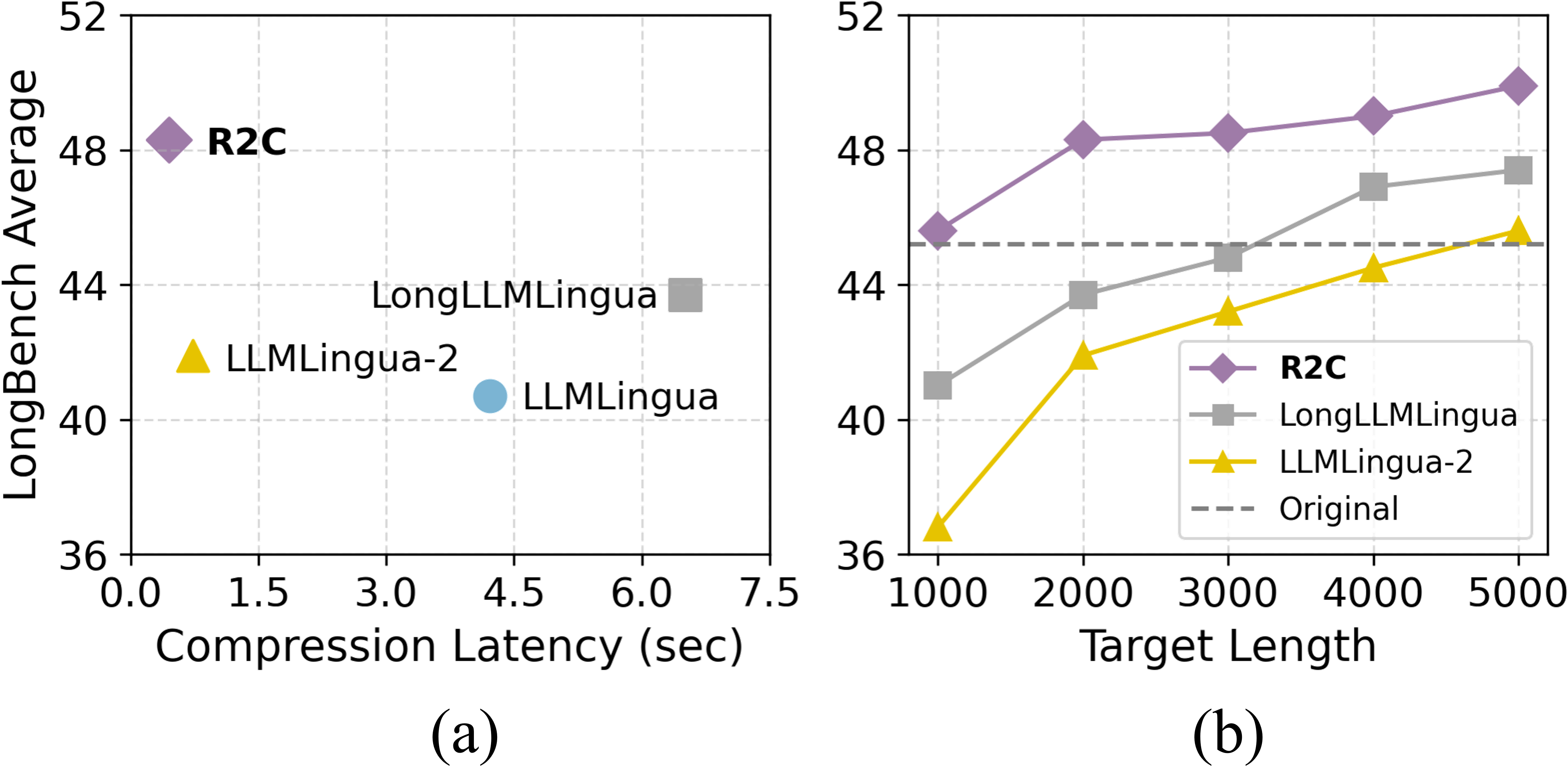}
\caption{Performance of GPT-3.5 with various compression methods in LongBench. (a): compression effectiveness-efficiency comparison. (b): effectiveness over varying compression ratios (2x--10x).}
\label{fig:fig_latency_targetlength}
\end{figure}

%% file: Tables/tab_e2e_efficiency.tex
\begin{table}[t]\small
\centering
\setlength{\tabcolsep}{4pt}
% \begin{adjustbox}{width=1\columnwidth,center}
\begin{tabular}{c|ccc|c}
\toprule
\begin{tabular}[c]{@{}c@{}}Compression\end{tabular}  & \begin{tabular}[c]{@{}c@{}}Comp.\\ latency\end{tabular} &  \begin{tabular}[c]{@{}c@{}}API\\ latency\end{tabular} & \begin{tabular}[c]{@{}c@{}}E2E\\ latency\end{tabular} & \begin{tabular}[c]{@{}c@{}}E2E latency\\ in \% \end{tabular} \\ \midrule
- & 0s                                                                                                     & 1.52s                                                  & 1.52s                                                  & 100.0\% \\ \midrule

R2C (5x) & 0.45s                                                                                                     & 0.88s                                                  & 1.33s                                                  & 87.5\% \\
R2C (10x) & 0.44s                                                                                                     & 0.68s                                                  & 1.11s                                                  & 74.0\% \\ \bottomrule
\end{tabular}
% \end{adjustbox}
% \setlength{\tabcolsep}{4pt}
\caption{End-to-end efficiency of R2C on LongBench dataset. Comp. latency indicates the latency for compressing prompts. E2E denotes the latency from the prompt compression to the black-box API (GPT-3.5). Note that latency is in seconds.}
\label{tab:e2e_efficiency}
\end{table}

%% file: Tables/tab_ablation.tex
\begin{table}[t]\small
\centering
% \begin{adjustbox}{width=1\columnwidth,center}
\begin{tabular}{c|c|c}
\toprule
Variants                   & NQ dev & \# tokens \\ \midrule
R2C                        & \textbf{58.44}   & 483       \\ \midrule
R2C w/ chunk only          & 57.01            & 485       \\
R2C w/ sentence only       & 57.58            & 480       \\
R2C w/ tokens only         & 51.86            & 478       \\ \midrule
T5-base initialize         & 44.88            & 483       \\
w/ decoder last layer only & 56.73            & 483       \\
unit aggregation: max      & 58.39            & 483       \\
unit aggregation: sum      & 57.58            & 482       \\ \bottomrule
\end{tabular}
% \end{adjustbox}
\caption{Ablation study of R2C on the NQ dev dataset. The results are based on LLaMA2-7B with 6x compressed prompts.}
\label{tab:ablation}
\end{table}

%% file: Figures_tex/fig_hyperparams.tex
\begin{figure}[t]
\centering
\includegraphics[width=0.48\textwidth]{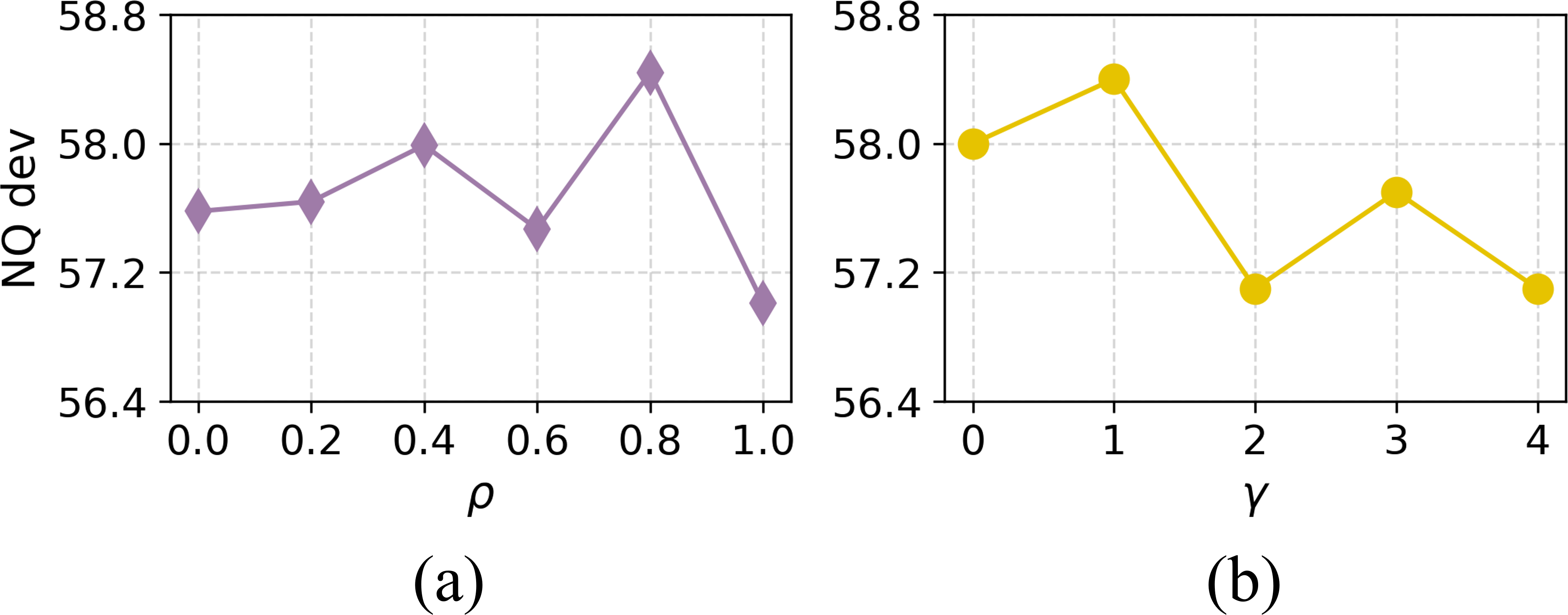}
\caption{Performance of LLaMA2-7B with R2C on the NQ dev dataset adjusting (a) the hierarchical ratio $\rho$ and (b) importance coefficient $\gamma$.}
\label{fig:fig_hyperparams}

\end{figure}

%% file: sec-conclusion.tex
\section{Conclusion}\label{sec:conclusion}

In this work, we explored the capabilities of multi-document readers for prompt compression and successfully bridged two different lines of research. We propose a novel prompt compression method, namely \textit{Reading To Compressing} (R2C), leveraging the Fusion-in-Decoder. R2C effectively captures global context and identifies salient information across multiple segments. Furthermore, by training the compressor using question-answering datasets, the most influential tokens are identified without using noisy pseudo-labels for compressed prompts. Experimental results demonstrate that R2C outperforms existing prompt compression methods by preserving semantic integrity and even surpasses uncompressed prompts by removing ambiguity.

%% file: sec-limitation.tex
\section{Limitations}\label{sec:limitation}
\noindent\textbf{Task Generalization}. While training QA captures important context across various benchmarks, it remains challenging to generalize the capability to all tasks. Prompt compression requires understanding both instructions and the entire prompt. Recent work~\cite{acl/YeBPRH23/FiD-ICL} validates the FiD structure in in-context learning and suggests excluding redundant questions. Given FiD's efficiency in capturing global context, further research, including diverse training strategies, remains to be explored.

\noindent\textbf{Dynamic Compression Ratios and Granularity}. R2C sets a target number of tokens for prompt compression, providing intuitive usability. However, each prompt has an optimal compression ratio, and using fewer tokens can remove noise and enhance performance in some cases. It is necessary to adjust the length of prompts dynamically based on each prompt. Additionally, the appropriate compression granularity, \eg, chunk, sentence, phrase, or token, should be adjustable. Even when compressing the same number of tokens, determining the appropriate granularity for each prompt and compressing accordingly can ensure that the most relevant information is retained, potentially improving performance across various tasks.

% 1. QA 학습이 다양한 벤치마크에서 중요한 부분을 잘 포착할 수 있는 것을 확인했지만 모든 task를 cover하기는 어려움. 다양한 prompt를 압축하기 위해서는 instruction을 이해할 뿐만 아니라 프롬프트를 전체적으로 이해하고 중요한 정보만을 남길 수 있어야 함. 최근 ~\cite{acl/YeBPRH23/FiD-ICL}은 FiD 구조가 In-Context learning에서도 적합함을 보이고, 특히 encoder에서 중복적으로 질문을 넣어주는 것을 제외하는 것을 제안하였음. FiD 구조가 매우 효율적으로 global context를 포착할 수 있는 구조인만큼, 더욱 다양한 학습을 통해 ???

% 2. R2C는 prompt를 압축하기 위해 target number of tokens를 설정함. 이는 사용자에게 가장 직관적인 활용성을 제공할 수 있는 방법이나, prompt 마다 적정 압축 비율이 존재하며 일부 task에서는 더 적은 토큰을 활용할 때 오히려 noise를 제거하여 성능이 향상하는 결과를 보임. prompt마다 dynamic하게 적절한 길이로 줄이는 시도가 필요함. 또한 prompt에 따라 적절한 granularity를 조절할 수 있어야함. 동일한 토큰 수를 압축하더라도 어떤 granularity를 활용하여 줄이는 것이 적절한지를 판단하고, 이에 맞게 압축한다면...

%% file: sec-ethics.tex
\section*{Ethics Statement}

This work fully respects ACL's ethical guidelines. We have utilized scientific resources available for research under liberal licenses, and our use of these tools is consistent with their intended applications.

% AI대학원, 나비효과, 오픈도메인
\section*{Acknowledgments} 
This work was supported by Institute of Information \& communications Technology Planning \& Evaluation (IITP) grant funded by the Korea government (MSIT) (No. RS-2019-II190421, RS-2022-II220680, RS-2022-II221045)

%% file: sec-appendix.tex
\newpage
\appendix
\input{Tables/tab_longbench_varying_t}

\input{Tables/tab_dataset_stat_longbench}

\section{Additional Experimental Setup}\label{sec:app_setup}
\noindent

\subsection{Dataset}\label{sec:app_dataset}
\noindent
\textbf{Natural Questions}~\cite{tacl/KwiatkowskiPRCP19/NQ}. The dataset contains Wikipedia articles and questions corresponding to Google search queries. We split the dataset into train, dev, and test sets following \citet{eacl/IzacardG21/FiD}. We further sample 1,757 dev samples for experiments, as shown in Table~\ref{tab:statistic}.

\noindent
\textbf{LongBench}~\cite{bai2023longbench}. The dataset is a multi-task benchmark for long context comprising 21 datasets across 6 task categories in both English and Chinese. The task types cover essential long-text application scenarios including single-document QA, multi-document QA, summarization, few-shot learning, code completion, and synthetic tasks. We examine the effectiveness of the proposed method only on \textit{English real-world datasets}, excluding Chinese datasets and synthetic tasks, as listed in Table~\ref{tab:statistic}.

\noindent
\subsection{Evaluation Metrics}\label{sec:app_metric}
\noindent
\noindent
\textbf{Natural Questions}. We adopt a Span Exact Match (Span EM) for an evaluation metric, aligned with previous works~\cite{emnlp/LesterAC21/softprompts, corr/abs-2307-03172/lost-in-the-middle, corr/abs-2310-06839/LongLLMLingua}. Span EM measures whether the answer is part of the generated answer of the target LLM, effectively reflecting the performance of LLMs on QA datasets rather than mere Exact Match (EM).
% Note that the official metric for Natural Questions is Exact Match (EM), which evaluates whether they exactly match the correct answer, but this is not suitable for language models~\cite{corr/abs-2307-03172/lost-in-the-middle} Instead, we adopt Span EM, which evaluates whether the generated text contains the answer.

\noindent
\textbf{LongBench}. The metric of each dataset is shown in Table~\ref{tab:statistic}. We follow the official metrics which are consistent with the original work. For NarrativeQA, Qasper, MultiFieldQA, HotpotQA, 2WikiMultihopQA, MuSiQue, and TriviaQA, we adopt the F1 score as a metric. We use the Rouge-L score~\cite{lin-2004-rouge} for GovReport, QMSum, MultiNews, and SAMSum, which are widely adopted in summarization tasks. For the TREC dataset, the classification accuracy is measured. Lastly, Edit Sim~\cite{SvyatkovskiyDFS20EditSim} (Levenshetein distance) is used for LCC and RepoBench-P, which is popularly used in code generation evaluation.

\noindent
\subsection{Evaluation Prompts}\label{sec:app_prompt}
\input{Tables/tab_prompt}

\noindent
Table~\ref{tab:prompt} describes the evaluation prompts used for generating answers with target LLMs. We follow the prompt of \citet{corr/abs-2307-03172/lost-in-the-middle} and \citet{bai2023longbench} for Natural Questions and LongBench, respectively.

\noindent
\section{Case Study}\label{sec:app_case_study}
% , fig:case_singledoc, fig:case_multidoc, fig:case_code
\noindent Figure~\ref{fig:case_nq},~\ref{fig:case_singledoc},~\ref{fig:case_multidoc},~\ref{fig:case_fewshot},~\ref{fig:case_summ}, and~\ref{fig:case_code} showcase case studies of how R2C effectively compresses lengthy prompts. We set the target length $T$ to 500 and 1,000 tokens for the Natural Questions and LongBench datasets, respectively. As demonstrated in the case studies, R2C successfully captures and retains the essential information from long inputs. By employing chunk- and sentence-level compression, R2C preserves the semantic integrity of the prompts, enabling the target LLM to better understand and respond to the prompts. These examples highlight the ability of R2C to efficiently compress prompts without sacrificing crucial details necessary for generating correct answers.

Natural Questions and the SingleDoc task require finding the specific part of a document relevant to the question from multiple documents or a very long source document. In Figure~\ref{fig:case_nq}, R2C gives a high score to "\textit{Linda Davis}" out of 20 documents and finds the part that says she sang "\textit{Does He Love You}" from the document. Also, in a long document of 9,845 tokens (Figure~\ref{fig:case_singledoc}), R2C gives a high score to the sentence that he worked in "\textit{3D printing and software development}" in "\textit{Tennessee}".

The MultiDoc task requires referencing multiple parts to generate accurate responses, making it difficult to compress due to the distribution of essential information. In Figure~\ref{fig:case_multidoc}, R2C captures the global context and retains documents that contain key information, such as references to "\textit{Night of Dark Shadows}" and "\textit{alcohol}".

The Summarization task requires generating a short summary from a long original text. Figure~\ref{fig:case_summ} shows an example from the GovReport dataset, generating a one-page summary from a 3,510 token long report. We can see that R2C preserves important information, such as the “\textit{eligibility}” and “\textit{coverage}” of the PSOB.

The FewShot Task provides several relevant demonstrations to the current question and asks the LLM to generate an answer by referring to them. Figure~\ref{fig:case_fewshot} illustrates the task of summarizing the dialogue in which “\textit{Jones and Angelina}” make an "\textit{appointment}". In this example, R2C gives high scores to conversations that remind “\textit{Audrey and Tom}” of a previous "\textit{appointment}."

The Code Completion task predicts the next line of the currently given code. To solve this task, it is necessary to reference multiple code files relevant to the current code. In Figure~\ref{fig:case_code}, the task is predicting the next line of "\texttt{owningAccount = getEucalyptusAccount() ;}", and we can observe that R2C gives a high sentence-level score to “\texttt{owningAccount = getEucalyptusAccount();}” and “\texttt{adminUser = getEucalyptusAdmin();}” among the existing code.

% In the Natural Questions shown in Figure~\ref{fig:case_nq}, the target length $T$ is set to 500 tokens, while for LongBench datasets, we show the compressed results with $T=1000$. 

\input{Figures_tex/case_study_nq}
\input{Figures_tex/case_study_singledoc}
\input{Figures_tex/case_study_multidoc}
\input{Figures_tex/case_study_summarization}
\input{Figures_tex/case_study_fewshot}
\input{Figures_tex/case_study_code}

%% file: Tables/tab_longbench_varying_t.tex
\begin{table*}[t]\small
\centering
\begin{tabular}{c|c|c|ccccc|c|c}
\toprule
Target LLM               & Compression          & $T$ & SingleDoc     & MultiDoc      & Summ.         & FewShot       & Code          & Avg.          & \# toks \\ \midrule
\multirow{6}{*}{GPT-3.5} & Original             & -               & 43.2          & 46.1          & \textbf{25.2} & 69.2          & \textbf{64.4} & 49.6          & 9,881     \\ \cmidrule{2-10} 
                         & \multirow{5}{*}{R2C} & 1,000           & 41.8          & 45.4          & 24.1          & 64.9          & 51.9          & 45.6          & 1,048     \\
                         &                      & 2,000           & 43.5          & \textbf{48.7} & 24.9          & 66.9          & 57.6          & 48.3          & 1,976     \\
                         &                      & 3,000           & 43.7          & 46.9          & 24.9          & 67.1          & 59.8          & 48.5          & 2,832     \\
                         &                      & 4,000           & 43.6          & 47.5          & 25.0          & 68.3          & 60.9          & 49.0          & 3,603     \\
                         &                      & 5,000           & \textbf{44.4} & 47.8          & 25.1          & \textbf{69.9} & 62.5          & \textbf{49.9} & 4,305     \\ \bottomrule
\end{tabular}
\caption{Performance of the GPT-3.5 on the LongBench benchmark with R2C, varying the target tokens $T$. \# tokens denotes the average number of tokens across all datasets based on the ChatGPT tokenizer. Since some prompts with lengths shorter than the target tokens are not compressed, the average number of tokens may be less than the target tokens $T$. The best performance is marked in \textbf{bold}.}
\label{tab:longbench_varying_t}
\end{table*}

%% file: Tables/tab_dataset_stat_longbench.tex
\begin{table*}[t]
\centering
\begin{tabular}{ccccc}
\toprule
Dataset & Source & Avg len & Metric & \# samples \\ \midrule
\multicolumn{5}{l}{\textit{Question Answering}} \\
Natural Questions (NQ) & Wikipedia & 3,018 & Span EM & 79,168/8,757/3,610  \\
\midrule

\multicolumn{5}{l}{\textit{Single-document QA (SingleDoc)}} \\
NarrativeQA & Literature, Film & 29,872 & F1 & 200 \\
Qasper & Science & 5,090 & F1 & 200 \\
MultiFieldQA-en & Multi-field & 6,988 & F1 & 150 \\ \midrule
\multicolumn{5}{l}{\textit{Multi-document QA (MultiDoc)}} \\
HotpotQA & Wikipedia & 12,867 & F1 & 200 \\
2WikiMultihopQA & Wikipedia & 7,188 & F1 & 200 \\
MuSiQue & Wikipedia & 15,650 & F1 & 200 \\ \midrule
\multicolumn{5}{l}{\textit{Summarization (Summ.)}} \\
GovReport & Government report & 10,276 & Rouge-L & 200 \\
QMSum & Meeting & 13,917 & Rouge-L & 200 \\
MultiNews & News & 2,642 & Rouge-L & 200 \\ \midrule
\multicolumn{5}{l}{\textit{Few-shot Learning (FewShot)}} \\
TREC & Web & 6,785 & Accuracy & 200 \\
TriviaQA & Wikipedia & 11,799 & F1 & 200 \\
SAMSum & Dialogue & 9,173 & Rouge-L & 200 \\ \midrule
\multicolumn{5}{l}{\textit{Code Completion (Code)}} \\
LCC & Github & 3,179 & Edit Sim & 500 \\
RepoBench-P & Github & 10,826 & Edit Sim & 500 \\ \bottomrule
\end{tabular}
\caption{Detailed statistics of Natural Questions~\cite{tacl/KwiatkowskiPRCP19/NQ} and LongBench~\cite{bai2023longbench}. For NQ, we split data into train, dev, and test set following \citet{eacl/IzacardG21/FiD}. For LongBench, we only include English datasets and omit a synthetic task for the evaluation in real-world settings. `Source' denotes the original source of the context. `Avg len' is the average token length of each dataset computed by the GPT-3.5 tokenizer. Note that we exclude the system message when counting the Avg len for the NQ dataset.}
\label{tab:statistic}
\end{table*}

%% file: Tables/tab_prompt.tex
\begin{table*}[ht!]\small
\centering
\resizebox{\linewidth}{!}{
\begin{tabular}{c|c|p{16cm}}
\toprule
\multicolumn{2}{c|}{\textbf{Dataset}} & \multicolumn{1}{c}{\textbf{Prompts}} \\ \midrule
\multicolumn{2}{c|}{Natural Questions} & <s>[INST] <<SYS>> You are a helpful, respectful and honest assistant. Always answer as helpfully as possible, while being safe. Please ensure that your responses are socially unbiased and positive in nature. If a question does not make any sense, or is not factually coherent, explain why instead of answering something not correct. If you don't know the answer to a question, please don't share false information. <</SYS>> \newline Write a high-quality answer for the given question using only the provided search results (some of which might be irrelevant). \newline \texttt{\{compressed\_context\}} \newline Question: \texttt{\{question\}} \newline Answer: [/INST] \\ \midrule

\multirow{2}{*}{LongBench} & \multirow{2}{*}{\begin{tabular}[c]{@{}c@{}}NarrativeQA \\ (SingleDoc)\end{tabular}} 
& You are given a story, which can be either a novel or a movie script, and a question. Answer the question as concisely as you can, using a single phrase if possible. Do not provide any explanation. \newline Story: \texttt{\{compressed\_context\}} \newline Now, answer the question based on the story as concisely as you can, using a single phrase if possible. Do not provide any explanation. \newline Question: \texttt{\{question\}} \newline Answer: \\ \cmidrule{2-3}

& \multirow{2}{*}{\begin{tabular}[c]{@{}c@{}}Qasper \\ (SingleDoc)\end{tabular}} 
& You are given a scientific article and a question. Answer the question as concisely as you can, using a single phrase or sentence if possible. If the question cannot be answered based on the information in the article, write "unanswerable". If the question is a yes/no question, answer "yes", "no", or "unanswerable". Do not provide any explanation. \newline Article: \texttt{\{compressed\_context\}} \newline Answer the question based on the above article as concisely as you can, using a single phrase or sentence if possible. If the question cannot be answered based on the information in the article, write "unanswerable". If the question is a yes/no question, answer "yes", "no", or "unanswerable". Do not provide any explanation. \newline Question: \texttt{\{question\}} \newline Answer: \\ \cmidrule{2-3}

& \multirow{2}{*}{\begin{tabular}[c]{@{}c@{}}MultiFieldQA-en \\ (SingleDoc)\end{tabular}} 
& Read the following text and answer briefly. \newline \texttt{\{compressed\_context\}} \newline Now, answer the following question based on the above text, only give me the answer and do not output any other words. \newline Question: \texttt{\{question\}} \newline Answer: \\ \cmidrule{2-3}

& \multirow{2}{*}{\begin{tabular}[c]{@{}c@{}} HotpotQA \\ (MultiDoc)\end{tabular}} 
& Answer the question based on the given passages. Only give me the answer and do not output any other words. \newline The following are given passages. \newline \texttt{\{compressed\_context\}} \newline Answer the question based on the given passages. Only give me the answer and do not output any other words. \newline Question: \texttt{\{question\}} \newline Answer: \\ \cmidrule{2-3}

& \multirow{2}{*}{\begin{tabular}[c]{@{}c@{}} 2WikiMultihopQA \\ (MultiDoc)\end{tabular}} 
& Answer the question based on the given passages. Only give me the answer and do not output any other words. \newline The following are given passages. \newline \texttt{\{compressed\_context\}} \newline Answer the question based on the given passages. Only give me the answer and do not output any other words. \newline Question: \texttt{\{question\}} \newline Answer: \\ \cmidrule{2-3}

& \multirow{2}{*}{\begin{tabular}[c]{@{}c@{}} MuSiQue \\ (MultiDoc)\end{tabular}} 
& Answer the question based on the given passages. Only give me the answer and do not output any other words. \newline The following are given passages. \newline \texttt{\{compressed\_context\}} \newline Answer the question based on the given passages. Only give me the answer and do not output any other words. \newline Question: \texttt{\{question\}} \newline Answer: \\ \cmidrule{2-3}\

& \multirow{2}{*}{\begin{tabular}[c]{@{}c@{}} GovReport \\ (Summ.) \end{tabular}} 
& You are given a report by a government agency. Write a one-page summary of the report. \newline Report: \newline \texttt{\{compressed\_context\}} \newline Now, write a one-page summary of the report. \newline Summary: \\ \cmidrule{2-3}

& \multirow{2}{*}{\begin{tabular}[c]{@{}c@{}} QMSum \\ (Summ.) \end{tabular}} 
& You are given a meeting transcript and a query containing a question or instruction. Answer the query in one or more sentences. \newline Transcript: \texttt{\{compressed\_context\}} \newline Now, answer the query based on the above meeting transcript in one or more sentences. \newline Query: \texttt{\{question\}} \newline Answer: \\ \cmidrule{2-3}

& \multirow{2}{*}{\begin{tabular}[c]{@{}c@{}} MultiNews \\ (Summ.) \end{tabular}} 
& You are given several news passages. Write a one-page summary of all news. \newline News: \texttt{\{compressed\_context\}} \newline Now, write a one-page summary of all the news. \newline Summary: \\ \cmidrule{2-3}

& \multirow{2}{*}{\begin{tabular}[c]{@{}c@{}} TREC \\ (FewShot) \end{tabular}} 
& Please determine the type of the question below. Here are some examples of questions. \newline \texttt{\{compressed\_context\}} \newline \texttt{\{question\}} \\ \cmidrule{2-3}

& \multirow{2}{*}{\begin{tabular}[c]{@{}c@{}} TriviaQA \\ (FewShot) \end{tabular}} 
& Answer the question based on the given passage. Only give me the answer and do not output any other words. The following are some examples. \newline \texttt{\{compressed\_context\}} \newline \texttt{\{question\}} \\ \cmidrule{2-3}

& \multirow{2}{*}{\begin{tabular}[c]{@{}c@{}} SAMSum \\ (FewShot) \end{tabular}} 
& Summarize the dialogue into a few short sentences. The following are some examples. \newline \texttt{\{compressed\_context\}} \newline \texttt{\{question\}} \\ \cmidrule{2-3}

& \multirow{2}{*}{\begin{tabular}[c]{@{}c@{}} LCC \\ (Code) \end{tabular}} 
& Please complete the code given below. \newline \texttt{\{compressed\_context\}} Next line of code: \\ \cmidrule{2-3}

& \multirow{2}{*}{\begin{tabular}[c]{@{}c@{}} RepoBench-P \\ (Code) \end{tabular}} 
& Please complete the code given below. \newline \texttt{\{compressed\_context\}} \texttt{\{question\}} Next line of code: \\ 
\bottomrule
\end{tabular}
}
\caption{Prompts used for the target LLM on Natural Questions~\cite{tacl/KwiatkowskiPRCP19/NQ} and LongBench~\cite{bai2023longbench}, following \citet{corr/abs-2307-03172/lost-in-the-middle} and \citet{bai2023longbench}. \texttt{\{compressed\_context\}} and \texttt{\{question\}} denote the compressed context and question, respectively.}
\label{tab:prompt}
\end{table*}

%% file: Figures_tex/case_study_nq.tex
\begin{figure*}[htb]
\centering
\begin{tcolorbox}[colframe=black,colback=white!10!white,title=Case study of Natural Questions,fonttitle=\bfseries]

\textbf{R2C Compressed Prompt with $T=500$ (481 tokens, Original length: 3,038 tokens)}\\
<s>[INST] <<SYS>>

You are a helpful, respectful and honest assistant. Always answer as helpfully as possible, while being safe. Please ensure that your responses are socially unbiased and positive in nature. If a question does not make any sense, or is not factually coherent, explain why instead of answering something not correct. If you don't know the answer to a question, please don't share false information.

<</SYS>>

Write a high-quality answer for the given question using only the provided search results (some of which might be irrelevant).
 
\colorbox{instructionbg}{\parbox{\dimexpr\linewidth-2\fboxsep}{Document [2](Title: \hl{Linda Davis}) Linda Davis Linda Kaye Davis (born November 26, 1962) is an American country music singer. Before beginning a career as a solo artist, she had three minor country singles in the charts as one half of the duo Skip \& Linda. Her highest chart entry is \hl{"Does He Love You", her 1993 duet with Reba McEntire}, which reached number one on the "Billboard" country charts and won both singers the Grammy for Best Country Vocal Collaboration.

Document [1](Title: Does He Love You) Does He Love You "Does He Love You" is a song written by Sandy Knox and Billy Stritch, and recorded as a duet by American country music artists Reba McEntire and Linda Davis. It is one of country music's several songs about a love triangle.

Document [5](Title: Linda Davis)  Linda Davis Linda Kaye Davis (born November 26, 1962) is an American country music singer. Before beginning a career as a solo artist, she had three minor country singles in the charts as one half of the duo Skip \& Linda.

Document [4](Title: Greatest Hits Volume Two (Reba McEntire album)) (then a background singer in McEntire's road band), was the first single and turned out to be a smash. "Does He Love You" is the first of three duets featuring Reba and Linda Davis.

Document [10](Title: Does He Love You) on Patti LaBelle's album, "Flame". It then shows Linda as a professional actress at a movie premiere, then cuts back to Reba smashing the picture with a

Document [11](Title: Shoot for the Moon (album))  Tim DuBois, president of Arista's Nashville division, said that "we rushed it" after Davis won a Grammy Award for her duet wirh Reba McEntire on "Does He Love You".

Document [7](Title: Reba McEntire) Out of the ten tracks were two new singles: the first, "Does He Love You", was a duet with}}

\colorbox{questionbg}{\parbox{\dimexpr\linewidth-2\fboxsep}{Question: who sings does he love me with reba

Answer: [/INST]}}
% Answer:
% Question: What field does Danny work in in Tennessee?

% Answer:

\vspace{0.5cm}

\textbf{Prediction (Llama2-7b-chat-hf)}
\newline Based on the search results provided, the answer to the question "Who sings 'Does He Love Me' with Reba" is Linda Davis. According to Document [1](Title: Does He Love You), the song "Does He Love You" was written by Sandy Knox and Billy Stritch, and recorded as a duet by American country music artists Reba McEntire and Linda Davis. Therefore, Linda Davis is the singer who collaborates with Reba

\vspace{0.5cm}

\textbf{Ground truth}
\newline Linda Davis

\end{tcolorbox}
\caption{Case study on the Natural Questions development set. The number of tokens is calculated using a ChatGPT tokenizer excluding system messages. The \hl{purple colorbox} indicates the core information to generate answers.}\label{fig:case_nq}
\end{figure*}

%% file: Figures_tex/case_study_singledoc.tex
\begin{figure*}[htb]
\centering
\begin{tcolorbox}[colframe=black,colback=white!10!white,title=Case study of SingleDocQA task in LongBench,fonttitle=\bfseries]

\textbf{R2C Compressed Prompt with $T=1,000$ (1,012 tokens, Original length: 9,845 tokens)}\\
Read the following text and answer briefly.
 
\colorbox{instructionbg}{\parbox{\dimexpr\linewidth-2\fboxsep}{My Aspergers Child: COMMENTS \& QUESTIONS [for Feb., 2017]

I emailed you a while back and you mentioned that I could email when I needed to. But the attorney, even though he was just vaguely familiar with Aspergers, has been very good with Craig. He has the compassion and excellence that is needed here.

~~~~~~~~~$\vdots$

 It's been 2 years now and he will not accept his diagnosis. I've tried telling him that it's not a bad thing, that there's been many, many very successful

It would have been so much easier to mention to my adult son that I think (I know he does, but want to ease into the subject)

he has Asperger's when we were living together two years ago. \hl{He has since moved to Tennessee working in his field of interest}

\hl{which is 3-D printing and software development.} I am so happy for him that he has found his way into a job that he truly enjoys

even though he's socially isolated.

He's not diagnosed and does not know he has it. How I know is his classic symptoms being sensory issues (fabric feeling like sandpaper)

I wanted to let you know about a research opportunity for children, teens, and young adults with autism.

 to manuel labor type jobs (which is not something he enjoys but he did it anyway).

At 19 1/2 he left to serve a 2 year full-time mission for our church. He completed his mission successfully. (I don't think it was without some struggle, stress and depression, but he was able to pick himself up and move on from those times).

~~~~~~~~~$\vdots$

 I hope you’ll agree it shows that starting work in the industry takes dedication and skill and that becoming a game designer isn’t just a fly-by-night job!}}

Now, answer the following question based on the above text, only give me the answer and do not output any other words.

\colorbox{questionbg}{\parbox{\dimexpr\linewidth-2\fboxsep}{Question: What field does Danny work in in Tennessee?

Answer:}}
% Answer:
% Question: What field does Danny work in in Tennessee?

% Answer:

\vspace{0.5cm}

\textbf{Prediction (GPT-3.5-turbo-1106)}
\newline 3-D printing and software development.

\vspace{0.5cm}

\textbf{Ground truth}
\newline 3-D printing and software development.

\end{tcolorbox}
\caption{Case study on the MultiFieldQA\_en dataset (SingleDoc) in the LongBench benchmark. The \hl{purple colorbox} indicates the core information to generate answers.}\label{fig:case_singledoc}
\end{figure*}

%% file: Figures_tex/case_study_multidoc.tex
\begin{figure*}[htb]
\centering
\begin{tcolorbox}[colframe=black,colback=white!10!white,title=Case study of MultiDocQA task in LongBench,fonttitle=\bfseries]

\textbf{R2C Compressed Prompt with $T=1,000$ (986 tokens, Original length: 4,137 tokens)}\\
Answer the question based on the given passages. Only give me the answer and do not output any other words.
\newline The following are given passages.
 
\colorbox{instructionbg}{\parbox{\dimexpr\linewidth-2\fboxsep}{~~~~~~~~~$\vdots$

Using aseptic packaging equipment, products can be packed in aseptic packaging. Pasteurized or UHT treated products packed into this format can be "shelf-stable", requiring no refrigeration.

Passage 3:

Bagman (disambiguation)

A bagman or bag man is a collector of dirty money for organized crime.

The Bag Man, a 2014 film

Bag Man (podcast), a 2018 podcast about Spiro Agnew's 1973 bribery and corruption scandal

"Bagman" (Better Call Saul), an episode of the television series Better Call Saul

See also

Bağban (disambiguation)

Bag boy (disambiguation)

Bag lady (disambiguation)

(1945 film), a 1945 American film starring Fred Allen

Passage 7:
\hl{The Bag Man

Purcell, Robert De Niro, and Sticky Fingaz. The film premiered on February 28, 2014}, in New York and Los Angeles.

Plot

Brutal gangster Dragna recruits professional killer Jack to pick up a bag and wait for his arrival at a motel.  Dragna stresses that Jack is not to open the bag or allow anyone to view its contents under any circumstances.  Confused as to why

~~~~~~~~~$\vdots$

Passage 8:

Una prostituta al servizio del pubblico e in regola con le leggi dello stato

\hl{Una prostituta al servizio del pubblico e in regola con le leggi dello stato (literally "A prostitute serving the public and complying with the laws of the state", also known as Prostitution Italian Style) is a 1970}  Italian comedy-drama film written and directed by Italo Zingarelli.For her performance Giovanna Ralli won the Grolla d'oro for best actress.

~~~~~~~~~$\vdots$

}}

Now, answer the following question based on the above text, only give me the answer and do not output any other words.

\colorbox{questionbg}{\parbox{\dimexpr\linewidth-2\fboxsep}{Question: Which film came out first, Una Prostituta Al Servizio Del Pubblico E In Regola Con Le Leggi Dello Stato or The Bag Man?

Answer:}}

\vspace{0.5cm}

\textbf{Prediction (GPT-3.5-turbo-1106)}
\newline Una Prostituta Al Servizio Del Pubblico E In Regola Con Le Leggi Dello Stato

\vspace{0.5cm}

\textbf{Ground truth}
\newline Una Prostituta Al Servizio Del Pubblico E In Regola Con Le Leggi Dello Stato

\end{tcolorbox}
\caption{Case study on the 2WikiMultihopQA dataset (MultiDoc) in the LongBench benchmark. The \hl{purple colorbox} indicates the core information to generate answers.}\label{fig:case_multidoc}
\end{figure*}

%% file: Figures_tex/case_study_summarization.tex
\begin{figure*}[htb]
\centering
\begin{tcolorbox}[colframe=black,colback=white!10!white,title=Case study of Summarization task in LongBench,fonttitle=\bfseries]

\textbf{R2C Compressed Prompt with $T=1,000$ (1,007 tokens, Original length: 3,510 tokens)}\\
You are given a report by a government agency. Write a one-page summary of the report.

Report:

\colorbox{instructionbg}{\parbox{\dimexpr\linewidth-2\fboxsep}{$\cdots$ \hl{There is no minimum amount of time a person must have served to be eligible for benefits. To be eligible for PSOB benefits as a law enforcement officer, firefighter, or chaplain, a person must have served in a "public agency" in an official capacity, with or without compensation.} For the purposes of PSOB eligibility, a public agency is defined as the federal government and any department, agency, OB eligibility, the definition of firefighter includes both professional firefighters and persons serving as an "officially recognized or designated member of a legally organized volunteer fire department." A chaplain is eligible for PSOB benefits $\cdots$ \hl{To qualify for coverage under the PSOB program, a public safety officer's disability or death must have been the result of a personal injury.}  evidence that the injury was the result of the individual's status as a public safety officer, that injury is covered by the PSOB program. The lump-sum PSOB death and disability benefit for FY2019 is \$359,316. The benefit amount is adjusted annually to reflect changes in the cost of living using the annual percentage change in the Consumer Price Index for Urban Consumers (CPI-U) for the one-year period ending in the previous June. If a public safety officer receives a disability benefit and later dies from the same injury, the officer's survivors may not receive a PSOB death benefit. The payable benefit amount is  Court struck down the federal Defense of Marriage Act in United States v. Windsor , the legally married spouse of a public safety office may be of the same sex as the officer.  or mental disability. PSOB death benefits are paid to eligible survivors in the following order: 1. if the officer is survived by only a spouse, 100\% of the death benefits are payable to the spouse; 2. if the officer is survived by a spouse and children, 50\% of the death benefits are payable to the spouse and the remaining 50\% is distributed equally among the officer's children; 3. if the officer is survived by only children, the death benefits are equally distributed among the officer's children; 4. if the $\cdots$
}}

Now, write a one-page summary of the report.

\colorbox{questionbg}{\parbox{\dimexpr\linewidth-2\fboxsep}{Summary: }}

\vspace{0.5cm}

\textbf{Prediction (GPT-3.5-turbo-1106)}
\newline The report provides information on the Public Safety Officers' Benefits (PSOB) program, which provides benefits to public safety officers, including law enforcement officers, firefighters, chaplains, FEMA employees, and emergency medical services members, who are permanently disabled or die as a result of a personal injury sustained in the line of duty. The report outlines the eligibility criteria for PSOB benefits, $\cdots$

\vspace{0.5cm}

\textbf{Ground truth}
\newline The Public Safety Officers' Benefits (PSOB) program provides cash benefits to federal, state, and local law enforcement officers; firefighters; employees of emergency management agencies; and members of emergency medical services agencies who are killed or permanently and totally disabled as the result of personal injuries sustained in the line of duty. The Public Safety Officers' Educational Assistance (PSOEA) program, a component of the PSOB program, provides higher-education assistance to the children and spouses of public safety officers killed or permanently disabled in the line of duty $\cdots$

\end{tcolorbox}
\caption{Case study on the GovReport dataset (Summarization) in the LongBench benchmark. The \hl{purple colorbox} indicates the core information to generate answers.}\label{fig:case_summ}
\end{figure*}

%% file: Figures_tex/case_study_fewshot.tex
\begin{figure*}[htb]
\centering
\begin{tcolorbox}[colframe=black,colback=white!10!white,title=Case study of FewShot task in LongBench,fonttitle=\bfseries]

\textbf{R2C Compressed Prompt with $T=1,000$ (1,034 tokens, Original length: 15,836 tokens)}\\
Summarize the dialogue into a few short sentences. The following are some examples.
 
\colorbox{instructionbg}{\parbox{\dimexpr\linewidth-2\fboxsep}{
~~~~~~~~~$\vdots$

Maisy: im hiding from a strange dinosaur :P wat happened to rach and cameron?

Steph: they split :/ yesterday

Maisy: oh no!

Dialogue: Tom: How are you today?

Tom: Great, I was afraid it could be difficult after last night

Summary: Jenny and Mia are already at the office.

Dialogue: Teresa: where should we meet?

\hl{Dialogue: Audrey: Honey, remember that we are going to buy a Christmas tree just after work.

Tom: I know, honey :)

Audrey: And a few gifts for kids

Tom: Do you have any ideas?

Audrey: Maybe some cars for Sam and some puzzles for Lucy?

Tom: It's a good idea.

Audrey: :*

Summary: Audrey and Tom will buy a Christmas tree and gifts for children after work.}

Dean: ?

Summary: Tom had a job interview.

It's been planned a long time ago.

Bill: I'm terribly sorry but would it be very incovenient to you if we postponed our meeting? I admit it's a little disappointing. These are some private issues.

Liam: which book?

~~~~~~~~~$\vdots$}}

\colorbox{questionbg}{\parbox{\dimexpr\linewidth-2\fboxsep}{Dialogue: Jones: Hey.
\newline Angelina: Hey.
\newline Angelina: Long time. How are you doing?
\newline Jones: I'm fine
\newline Jones: You?
\newline Angelina: I'm cool too.
\newline Jones: You think we can meet today later in the afternoon in town?
\newline Angelina: Definitely.
\newline Jones: Okay. I will call you to confirm where we will meet.
\newline Angelina: Cool
\newline Summary: }}

\vspace{0.5cm}

\textbf{Prediction (GPT-3.5-turbo-1106)}
\newline Jones and Angelina plan to meet in town later in the afternoon.

\vspace{0.5cm}

\textbf{Ground truth}
\newline Jones and Angelina will meet in town in the afternoon.

\end{tcolorbox}
\caption{Case study on the SAMSum dataset (FewShot) in the LongBench benchmark. The \hl{purple colorbox} indicates the core information to generate answers.}\label{fig:case_fewshot}
\end{figure*}

%% file: Figures_tex/case_study_code.tex
\lstset{
  basicstyle=\ttfamily\small,
  tabsize=4, % Adjust the tab size as needed
  frame=single, % code 주변의 네모 박스
  backgroundcolor=\color{white},
  xleftmargin=0.5em,
  xrightmargin=0.5em,
  % numbers=left,
  % numberstyle=\tiny,
  breaklines=true,
  escapeinside={(*@}{@*)} % Allows LaTeX inside listings
}

\begin{figure*}[htb]
\centering
\begin{tcolorbox}[colframe=black,colback=white!10!white,title=Case study of Code Completion task in LongBench,fonttitle=\bfseries]

\textbf{R2C Compressed Prompt with $T=1,000$ (980 tokens, Original length: 9,403 tokens)}

Please complete the code given below. 
 
% \colorbox{instructionbg}{\parbox{\dimexpr\linewidth-2\fboxsep}{
% \newline\hspace{2cm}$\vdots$
% \newline\hspace{1cm}\}
% \newline\hspace{1cm}private static final ArrayList<? extends UpgradeTask> upgrades = Lists.newArrayList(Setup.INSTANCE, CopyBucketsToOSG.INSTANCE,}}

\begin{lstlisting}
    (*@ $\vdots$ @*)
              continue;
            } else {
              // Add it to the map
              accountIdAccountMap.put(grantInfo.getUserId(), account);
            }
          } catch (Exception e) { // In case the account is deleted, skip the grant
            LOG.warn("Account ID " + grantInfo.getUserId() + " does not not exist. Skipping this grant");
            deletedAccountIds.add(grantInfo.getUserId());
            continue;
          }
        }
   * This method transforms a Walrus bucket to an OSG bucket.
            owningAccount = getEucalyptusAccount();
            owningUser = getEucalyptusAdmin();
          } else if (noCanonicalIdAccountIds.contains(walrusBucket.getOwnerId())) { // If canonical ID is missing, use eucalyptus admin account
            LOG.warn("Account ID " + walrusBucket.getOwnerId() + " does not have a canonical ID. Changing the ownership of bucket "
                + walrusBucket.getBucketName() + " to eucalyptus admin account");
            (*@ \hl{owningAccount = getEucalyptusAccount();} @*)
                // This is to avoid insert null IDs into cached sets/maps
 This is executed in the {@code ModifyWalrusBuckets} stage</li>
   *
            (*@ \hl{adminUser = getEucalyptusAdmin();} @*)
            } else if (noCanonicalIdAccountIds.contains(walrusObject.getOwnerId())) { // If canonical ID is missing, use eucalyptus admin account
              LOG.warn("Account ID " + walrusObject.getOwnerId() + " does not have a canonical ID. Changing the ownership of object "
                  + walrusObject.getObjectKey() + " in bucket " + walrusObject.getBucketName() + " to eucalyptus admin account");
              owningAccount = getEucalyptusAccount();
                  owningAccount = getEucalyptusAccount();

\end{lstlisting}

Next line of code:

\vspace{0.5cm}

\textbf{Prediction (GPT-3.5-turbo-1106)}
\begin{lstlisting}
```
adminUser = getEucalyptusAdmin();
```
\end{lstlisting}

\vspace{0.5cm}

\textbf{Ground truth}
\begin{lstlisting}
                  adminUser = getEucalyptusAdmin();
\end{lstlisting}

\end{tcolorbox}
\caption{Case study on the LCC dataset (Code) in the LongBench benchmark. The \hl{purple colorbox} indicates the core information to generate answers.}\label{fig:case_code}
\end{figure*}